\documentclass[runningheads,orivec]{llncs}

\usepackage[T1]{fontenc}
\usepackage{graphicx}
\usepackage{hyperref}
\usepackage{color}

\hypersetup{
    colorlinks=true,
    linkcolor=blue,
    filecolor=magenta,
    urlcolor=cyan,
    citecolor=blue, 
    pdftitle={On the Domain Robustness of Contrastive Vision-Language Models}
    pdfauthor={Mario Koddenbrock},
}

\usepackage{bbm}
\usepackage{amsmath,amssymb,amsfonts}
\usepackage{xcolor}
\usepackage{textcomp}
\usepackage{manyfoot}
\usepackage{booktabs}
\usepackage{listings}
\usepackage{bm}
\usepackage{todonotes}
\usepackage{makecell}
\usepackage{pifont}

\begin{document}
\title{On the Domain Robustness of Contrastive Vision-Language Models}
\titlerunning{Domain Robustness of Vision-Language Models}

\author{
Mario Koddenbrock\orcidID{0000-0003-3327-7404} \and
Rudolf Hoffmann\orcidID{0009-0006-5363-0278} \and
David Brodmann\orcidID{0009-0007-7664-4389} \and
Erik Rodner\orcidID{0000-0002-3711-1498}
}

\authorrunning{Koddenbrock et al.}

\institute{
KI-Werkstatt/Fachbereich 2, 
University of Applied Sciences Berlin,\\ 
Wilhelminenhofstr. 75A, 12459 Berlin, Germany\\
\email{firstname.lastname@htw-berlin.de}
}

\newcommand\condon{\;|\;}
\newcommand\deepbench{\texttt{DeepBench}}
\newcommand\clip{\texttt{CLIP}}
\newcommand\siglip{\texttt{SigLIP}}
\newcommand\alignmodel{\texttt{ALIGN}}
\newcommand\llava{\texttt{LLaVA}}
\newcommand\gemma{\texttt{Gemma 3}}

\definecolor{customgreen}{HTML}{9ADE7B}
\definecolor{customred}{HTML}{FF8F8F}

\maketitle

\begin{abstract}

In real-world vision-language applications, practitioners increasingly rely on large, pretrained foundation models rather than custom-built solutions, despite limited transparency regarding their training data and processes. 
While these models achieve impressive performance on general benchmarks, their effectiveness can decline notably under specialized domain shifts, such as unique imaging conditions or environmental variations. 
In this work, we introduce \deepbench{}, a framework designed to assess domain-specific robustness of vision-language models (VLMs).  
\deepbench{} leverages a large language model (LLM) to generate realistic, context-aware image corruptions tailored to specific deployment domains without requiring labeled data. 
We evaluate a range of contrastive vision-language architectures and architectural variants across six real-world domains and observe substantial variability in robustness, highlighting the need for targeted, domain-aware evaluation. 
\deepbench{} is released as open-source software\footnote{\deepbench{} is available at \url{https://github.com/ml-lab-htw/deepbench}.} to support further research into domain-aware robustness assessment.

\end{abstract}


\section{Introduction}\label{sec:intro}
Ensuring robust performance in computer vision applications has been a longstanding challenge, with numerous benchmarks and evaluation strategies proposed for common corruptions and adversarial attacks~\cite{hendrycks2019robustness,taori2020measuring,yin2019fourier,croce2020robustbench}. While such work has significantly advanced our understanding of robustness under controlled conditions, it often overlooks the \emph{domain-specific} factors that arise in real-world scenarios. Consequently, even high-performing models can fail when facing new contexts—such as specialized manufacturing processes or unique medical imaging conditions—that diverge from the distribution seen during training~\cite{wang2024sober}.

This challenge becomes increasingly important with the growing adoption of large VLMs trained on massive, weakly curated datasets scraped from the internet~\cite{bommasani2021opportunities}. Although these models excel at generic tasks (e.g., zero-shot image classification and retrieval), they may exhibit reduced reliability on domains where object appearance, environmental variability, or imaging protocols differ sharply from the data seen during pretraining~\cite{zheng2023preventing}. 
Notably, many VLMs expose little information about the exact nature of their training data or data augmentation strategies, making it difficult to anticipate whether they will maintain accuracy and consistency when deployed in unfamiliar environments.

\begin{figure}[tbp]
    \centering
    \includegraphics[width=\linewidth, keepaspectratio, 
    alt={A diagram illustrating the DeepBench framework workflow in several stages. The process begins with two inputs: a set of predefined corruption methods and a domain-specific description. These are fed into a Large Language Model (LLM), which in turn generates a tailored set of image corruptions. These corruptions are then applied to an image dataset. A vision-language model is evaluated on this corrupted dataset, and the final stage shows the results being stored and visualized through an interactive frontend.}]
    {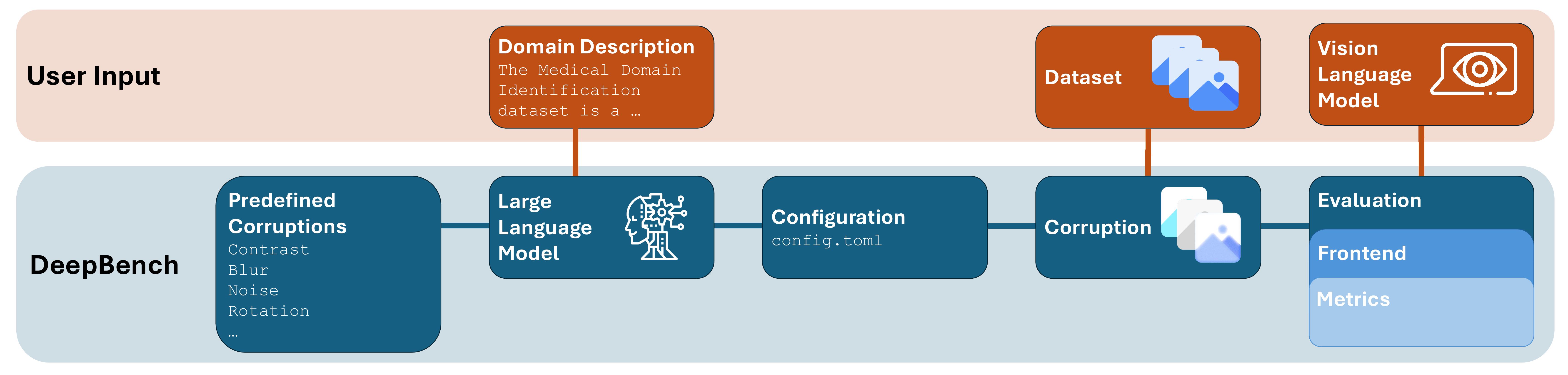}
    \caption[Overview of the \deepbench framework]{Overview of the \deepbench framework. A predefined set of corruption methods and a domain-specific description are used to prompt a Large Language Model (LLM), which outputs a tailored set of image corruptions. These are applied to a dataset, and model performance is evaluated under various perturbations. Results are stored and visualized through an interactive frontend}
    \label{fig:deepbench_overview}
\end{figure}

To tackle this gap, we introduce \deepbench{} (Fig.~\ref{fig:deepbench_overview}), a general-purpose framework that evaluates domain-specific robustness \emph{without} requiring labeled data. At its core \deepbench{} is an LLM-guided corruption generation pipeline: given a high-level domain description (such as imaging conditions or expected variability), \deepbench{} selects tailored perturbations from a predefined set to mimic real-world challenges. These corruptions, combined with an unsupervised consistency metric called \emph{label flip probability}, facilitate a thorough robustness assessment. By quantifying how often a model changes its prediction under perturbation, label flip probability offers a practical way to evaluate robustness even when ground-truth labels are scarce or expensive to obtain.

We focus on three contrastive VLMs—\clip~\cite{radford2021learning}, \siglip~\cite{zhai2023sigmoid}, and \alignmodel~\cite{jia2021scaling}, each pretrained on large-scale image-text pairs. Our evaluation covers six real-world domains, ranging from everyday photography to industrial and medical applications. 
In addition to standard accuracy-based metrics, we assess prediction consistency under corruption using label flip probability. We also analyze architectural variants of \clip{}, finding that robustness improves with larger model capacity, finer patch resolution, and the use of QuickGELU activations—though even the strongest models exhibit domain-specific weaknesses.

The \deepbench{} framework is open-source and designed to integrate with custom datasets or workflows. Its modular structure comprises a configurable backend for corruption generation and metric computation, alongside a user-facing interface for browsing and visualizing results. By emphasizing domain realism and data-scarce conditions, \deepbench{} highlights robustness failure modes that more generic benchmarks miss~\cite{koh2021wilds}.

\noindent
Our main contributions are:
\begin{enumerate}
    \item \textbf{An open-source framework for domain-specific robustness evaluation.} \deepbench{} enables unsupervised robustness benchmarking using LLM-guided corruption generation tailored to deployment scenarios.

    \item \textbf{Validation of LLM-guided corruption generation.} We demonstrate that LLM-generated corruption strategies are both consistent and context-aware across diverse domains, enabling effective simulation of real-world robustness challenges.

    \item \textbf{A systematic analysis of contrastive vision-language models under domain shift.} We benchmark leading contrastive VLMs across six real-world domains, showing that \clip{} achieves the highest overall robustness, while performance varies substantially by task and corruption type.

    \item \textbf{Insights into architectural design for robustness.} We analyze architectural variants and show that model capacity, patch size, input resolution, and activation function all influence robustness—highlighting \clip{} ViT-L/14 with QuickGELU as the most resilient configuration.
\end{enumerate}

\section{Related Work}

\paragraph{General Robustness in Computer Vision:}
Robustness has long been a central topic in computer vision, driven by the need to ensure stable performance under real-world variability. Efforts to measure and improve robustness against common corruptions and adversarial attacks date back to benchmarks like ImageNet-C and ImageNet-P~\cite{hendrycks2019robustness}, as well as work on adversarial examples~\cite{goodfellow2015explaining}. Later studies revealed that even mild natural distribution shifts can substantially degrade performance~\cite{taori2020measuring}. While these efforts provide valuable insights, they largely focus on \emph{general-purpose} corruptions, overlooking how domain-specific factors can reduce model reliability in real-world applications.

\paragraph{Robust Learning Approaches:}
Improving robustness often involves modifying the training process, for instance via adversarial training~\cite{madry2018towards}, distributionally robust optimization (DRO)~\cite{duchi2019distributionally}, or advanced data augmentation~\cite{cubuk2019autoaugment,lim2019fast,yin2019fourier}. Such methods typically aim to harden models against diverse perturbations. However, they are no guarantee for resilience and should therefore be assessed post-hoc, regardless of the training method.

\paragraph{Robustness in Vision-Language Models:}
VLMs such as \clip{}, \alignmodel{}, and \siglip{} have demonstrated strong zero-shot capabilities, leveraging large-scale image-text pretraining~\cite{fang2022data,zhai2023sigmoid,xu2023metaclip}. While open-source variants like \texttt{OpenCLIP}~\cite{radford2021learning} provide transparency, many pretrained datasets remain opaque~\cite{bommasani2021opportunities}, complicating bias and robustness analysis. Despite their scale, VLMs are prone to spurious correlations~\cite{wang2024sober} and often underperform in domain-specific settings. Existing solutions—such as fine-tuning~\cite{han2025alignclip}, ensembling~\cite{kar2024brave}, or test-time adaptation~\cite{maharana2024enhancing}—typically require labeled or generic data. Our work introduces a label-free pipeline to evaluate VLMs under realistic, domain-specific corruptions.

\paragraph{Domain-Specific Evaluations:}
Despite recent advances in general robustness methods, specialized application domains such as industrial inspection~\cite{bergmann2019mvtec} or scientific imaging~\cite{verma2024beyond} often present unique distribution shifts that generic techniques fail to capture~\cite{garg2023rlsbench}. In these contexts, labeled or source data may be scarce or inaccessible~\cite{li2022source}. Our framework addresses this by enabling realistic, zero-shot robustness assessments without requiring labels, providing practical insights into VLM performance under domain-specific shifts.

\section{\deepbench: A Framework for Domain-Specific Robustness Evaluation}

\subsection{Mathematical Framework and Robustness Metrics}
\label{par:metrics}

Using pretrained models without any fine-tuning is increasingly common, particularly for multimodal systems controlled solely through prompting. In classical computer vision, one typically considers a pretrained model \( f \) mapping the input space \( \mathcal{X} \) (e.g., images) to an output space \( \mathcal{Y} \) (e.g., a softmax over \( M \) classes with \( \mathcal{Y} = [0,1]^M \)). 
In the special case of pretrained contrastive VLMs, \( f \) consists of an image encoder \( \phi_I: \mathcal{X} \rightarrow \mathcal{Z} \) and a text encoder \( \phi_T: \mathcal{Y} \rightarrow \mathcal{Z}\) mapping to an shared embedding space \( \mathcal{Z} \subset \mathbb{R}^d \). 
\( f \) then performes prediction via some similarity measure  \( \mathrm{sim}: \mathcal{Z} \times \mathcal{Z} \rightarrow \mathbb{R} \)  (typically cosine similarity) between the image embedding of a sample \( \bm{x} \) and the embeddings of all possible labels:
\begin{align} \label{eq:y_hat}
f(\bm{x}) = \arg\max_{y \in \mathcal{Y}}\, \mathrm{sim}\Big(\phi_I(\bm{x}), \phi_T(y)\Big) \enspace.
\end{align}

An application domain \( \mathcal{D} \subseteq \mathcal{X} \) reflects the conditions under which images are acquired, inducing a domain-specific data distribution \( p(\bm{x}, y \mid \mathcal{D}) \). 
In the classical (supervised) evaluation, the overall performance of model \( f \) in \( \mathcal{D} \) is defined using a example-based 0–1 loss function $L$, such as the misclassification error \( L(z, z') = [z \neq z'] \):
\begin{align} \label{eq:perf_classic}
\varepsilon(f, \mathcal{D}) 
&= \mathbb{E}_{\bm{x}, y \sim p(\bm{x}, y \condon \mathcal{D})} \left( L(y, f(\bm{x})) \right).
\end{align}

Since the joint distribution \( p(\bm{x}, y \mid \mathcal{D}) \) is unknown, it is commonly (e.g. ~\cite{chapelle2000vicinal}) approximated by a test set \( \{ (\bm{x}_i, y_i)\}_{i=1}^{m} \) using Dirac delta functions:
\begin{align}
p(\bm{x}, y \mid \mathcal{D}) &= \frac{1}{m} \sum_{i=1}^{m} \delta(y - y_i)\,\delta(\bm{x} - \bm{x}_i) \enspace.
\end{align}

In contrast, vicinal learning~\cite{vapnik1999nature,vapnik1999overview}, replaces the Dirac delta functions with a smoothing function $\mathcal{K}$, which can, for example, consider a smooth region around each test example $\bm{x}_i$:
\begin{align}
p(\bm{x}, y \condon \mathcal{D}) &= \frac{1}{m} \sum\limits_{i=1}^m \delta(y - y_i) \left( \int \mathcal{K}(\bm{x}, \bm{x}_i) \;\mathrm{d}\bm{x} \right) \enspace.
\end{align}

The work of \cite{chapelle2000vicinal} showed that for learning with this scheme and a Gaussian kernel $\mathcal{K}$, classical Tikhonov regularization evolves. 
However, a Gaussian assumption often does not align with complex data, such as images, and it completely ignores additional domain knowledge that might be available. One type of domain knowledge that is natural to model involves invariant transformations $c(\cdot, \bm{\theta})$ of inputs that do not alter the output.

Therefore, instead of placing a Dirac delta at $\bm{x}_i$, it is more natural to marginalize over all possible transformation parameters $\bm{\theta}$ that do not lead to output changes and then approximate this marginalization with empirical samples of these parameters:
\begin{align}
\mathcal{K}(\bm{x}, \bm{x}_i) 
&= \mathbb{E}_{\bm{\mathbf{\theta}} \sim p(\bm{\theta} \condon \mathcal{D}) } \left(
\delta(\bm{x} - c(\bm{x}_i, \bm{\theta}))
\right)
\approx \frac{1}{K} \sum\limits_{k=1}^K \delta(\bm{x} - c(\bm{x}_i, \bm{\theta}_k)) \enspace,
\end{align}
where $(\bm{\theta}_k)_{k=1}^K$ are sampled from the domain-specific distribution $p(\bm{\theta} \condon \mathcal{D})$. 
Plugging this approximation into our performance measure from (\ref{eq:perf_classic}), we obtain the classical evaluation on a set of augmented test samples:

\begin{align} \label{eq:eps_1}
\varepsilon(f, \mathcal{D}) 
&\approx \frac{1}{mK}\sum_{i=1}^{m}\sum_{k=1}^{K} L(f(c(\bm{x}_i,\bm{\theta}_k)), y_i) 
= \frac{1}{mK}\sum_{i=1}^{m}\sum_{k=1}^{K} [ f(c(\bm{x}_i,\bm{\theta}_k)) \neq y_i ] 
\end{align}

\subsubsection{Extension to Unsupervised Robustness:}

In many real-world applications, ground-truth labels are unavailable. 
In such cases, robustness can be quantified by measuring the consistency of the model's predictions under corruption. 
Instead of using supervised loss \( L \), we consider whether the prediction changes when a corruption is applied, i.e., we replace the label \( y \) in Eq.~\ref{eq:perf_classic} with the clean prediction of image \( \bm{x} \). 
The \emph{label flip probability (FP)} of the model \( f \) is defined as  
\begin{align} \label{eq:fp}
\mathrm{FP}^f(\bm{x}) &= \mathbb{E}_{\bm{\theta} \sim p(\bm{\theta} \mid \mathcal{D})} \left[ f\big(c(\bm{x}, \bm{\theta})\big) \neq f(\bm{x}) \right] \enspace.
\end{align}

Transformed into the embedding space, a label flip means that applying a corruption \( c(\cdot, \bm{\theta}) \) shifts the image embedding such that there exists a label \( y' \neq f(\bm{x}) \) satisfying:
\begin{align}
\mathrm{sim}(\phi_I(c(\bm{x}, \bm{\theta})), \phi_T(y')) > \mathrm{sim}(\phi_I(c(\bm{x}, \bm{\theta})), \phi_T(f(\bm{x}))) \enspace.
\end{align}

To approximate the expectation from Eq.~\ref{eq:fp}, we simply evaluate on the sampled transformation parameters \( (\bm{\theta}_k)_{k=1}^K \):
\begin{align} \label{eq:fp_approx}
\mathrm{FP}^f(\bm{x}) \approx \frac{1}{K} \sum_{k=1}^K [ f\big(c(\bm{x}, \bm{\theta}_k)\big) \neq f(\bm{x}) ] \enspace
\end{align}

and derive a dataset-level flip rate \( \mathrm{FP}^f \), by averaging Eq.~\ref{eq:fp_approx} over all test samples. Finally, we get the \emph{mean flip rate} relative to an baseline model as
\begin{align}
\mathrm{mFR} &= \frac{\mathrm{FP}^f}{\mathrm{FP}^{\mathrm{baseline}}} \enspace.
\end{align}

\subsubsection{Mean and Relative Corruption Error (mCE, rCE)}
When ground-truth labels are available, we first establish a baseline using the \emph{clean balanced accuracy}—balanced accuracy computed on the unperturbed test set. 
Balanced accuracy, defined as the macro-average of per-class recall, compensates for class-imbalance bias.
Building on this baseline, we adopt the error-based robustness metrics introduced by Hendrycks et al.~\cite{hendrycks2019robustness}.

We consider \( b: \mathcal{X} \rightarrow \mathcal{Y} \) a baseline model and \( \tilde{\epsilon} (f, c, \mathcal{D}) \) the error rate of \(f\) on \(\mathcal{D}\), when corruption \(c\) is applied (averaged over all corresponding configurations).
Let \( \mathcal{C}  \) now be a set of different types of corruptions (i.e. blur, contrast, lightning etc.), then the \emph{mean Corruption Error (mCE)} is defined as
\begin{align}
\mathrm{mCE} &= \frac{1}{\lvert \mathcal{C} \rvert} \sum_{c \in \mathcal{C}} \frac{\tilde{\epsilon}(f, c, \mathcal{D})}{\tilde{\epsilon}(b, c, \mathcal{D})} \enspace.
\end{align}

Let further \( \tilde{\epsilon}(f, \mathcal{D}) \) be the clean error rate of \(f\) on \(\mathcal{D}\), then the \emph{relative Corruption Error (rCE)} is defined as:
\begin{align}
\mathrm{rCE} &= \frac{1}{\lvert \mathcal{C} \rvert} \sum_{c \in \mathcal{C}} \frac{\tilde{\epsilon}(f, c, \mathcal{D}) - \tilde{\epsilon}(f, \mathcal{D})}{\tilde{\epsilon}(b, c, \mathcal{D}) - \tilde{\epsilon}(b, \mathcal{D})} \enspace.
\end{align}
\subsection{Evaluation Framework}
\deepbench{} systematically evaluates model robustness across specialized domains using a modular architecture that combines automated corruption generation, metric-based evaluation, and interactive result visualization. At its core, the backend processes high-level domain descriptions and employs an LLM—default is \texttt{GPT-4o}~\cite{openai2023gpt4}—to generate a tailored set of corruptions from a predefined list. This list, detailed in Tables~\ref{tab:basic_corruption_methods} and~\ref{tab:advanced_corruption_methods}, includes corruption methods like geometric distortions, noise, color shifts, or simulated environmental effects. To guide the LLM’s reasoning, we use structured prompts consisting of a role definition, a domain description, a list of available corruptions with brief explanations, and explicit output format instructions (see \ref{app:prompts}, Listing~\ref{lst:prompting_example}). The generated output is parsed into a configuration file that specifies which corruptions to apply and with which parameters. The overall workflow is shown in Figure~\ref{fig:deepbench_overview}.

Each image in the selected dataset is processed under all defined corruption conditions, and model performance is measured using metrics such as balanced accuracy and label flip probability (see Section~\ref{par:metrics}). Results are stored in a MongoDB database and visualized through a browser-based frontend that supports interactive comparisons across models and domains. The framework is designed for extensibility: researchers can easily add new corruptions and configurations via a simple Python interface, enabling adaptation to emerging robustness challenges with minimal effort.

\subsection{Application Scenarios and Datasets} 
\label{par:applications}

\deepbench{} currently provides six diverse application scenarios, each representing a distinct real-world domain with its own imaging conditions, data distributions, and robustness challenges. 
Further application scenarios can be easily integrated into our open-source framework by specifying a high-level domain description (see Fig.~\ref{fig:deepbench_overview}), which informs the corruption generation pipeline. Providing details such as image source, expected variability, and deployment context, an LLM translates it into a tailored set of corruption transformations.

\begin{enumerate}
    \item \textbf{Medical Domain Identification:} 
    We use eight datasets covering diverse medical imaging types, including chest X-rays \cite{wang2017chestx}, hand X-rays for bone age assessment \cite{halabi2019rsna}, knee X-rays for osteoarthritis diagnosis \cite{sarhan2024knee}, dental X-rays \cite{ali2016detection}, ultrasound images for nerve segmentation \cite{baby2017automatic}, breast ultrasound images \cite{al2020dataset}, liver fibrosis ultrasound images \cite{joo2023classification}, and mammography images \cite{lee2017curated}. Each image is labeled according to its medical domain, with 126–999 samples per class.

    \item \textbf{Autonomous Driving (KITTI) \cite{geiger2012we}:} 
    KITTI (Karlsruhe Institute of Technology and Toyota Technological Institute) is an autonomous driving dataset, consisting of hours of traffic scenarios recorded with a variety of sensor modalities, including high-resolution RGB images. These images were cropped around non-overlapping object bounding boxes and re-labeled into four categories: \textit{Car}, \textit{Person}, \textit{Tram}, \textit{Truck}, with 156–4660 samples per class.

    \item \textbf{Manufacturing Quality Control (MVTec AD) \cite{bergmann2019mvtec}:} 
    The MVTec anomaly detection dataset (MVTec AD) is repurposed as a 15-class object classification task over industrial items, with 60–391 samples per class.

    \item \textbf{People and Face Recognition (FER13) \cite{goodfellow2013challenges}:} 
    The Facial Expression Recognition 2013 (FER13) dataset contains 35,887 grayscale 48×48 face images labeled with seven emotion categories, with 751–1093 samples per class.

    \item \textbf{Satellite and Aerial Imaging (AID) \cite{xia2017aid}:} 
    The Aerial Image Dataset (AID) offers 15 scene categories (e.g., farmland, airport) with 800 high-res RGB images per class from Google Earth.

    \item \textbf{Handheld Image Recognition (Food-101) \cite{bossard2014food}:} 
    Food-101 is a dataset including 101,000 images across 101 food categories (1,000 per class), showing a range of dishes under varied real-world conditions.
\end{enumerate}

\subsection{Built-In Vision Language Models}
\label{par:models}
\deepbench{} integrates with widely used APIs like Hugging Face~\cite{wolf2019huggingface} and OpenCLIP~\cite{cherti2023reproducible}, providing access to a broad range of pretrained vision-language models (VLMs). This study focuses on three representative models—\clip{}, \siglip{}, and \alignmodel{}—chosen for their architectural diversity and popularity. Unless noted otherwise, all models are evaluated at an input resolution of $224\times224$.

New models can be added via two mechanisms: (1) specifying the model name in the configuration file for API-based loading, or (2) implementing a lightweight Python wrapper for custom architectures. This modular design allows easy extension to new inference pipelines and VLM variants.

\begin{enumerate}
    \item \textbf{\clip}~\cite{radford2021learning} (OpenAI) is a contrastive model trained on 400M image-text pairs using separate Transformer encoders for vision and text. It aligns both modalities in a shared embedding space for zero-shot classification.

    \item \textbf{\siglip}~\cite{zhai2023sigmoid} (Google Research) builds on \clip{} but replaces the softmax loss with a sigmoid-based variant, improving performance in multi-label settings.

    \item \textbf{\alignmodel}~\cite{jia2021scaling} (Google Brain) is trained on over 1B noisy image-text pairs. It uses an EfficientNet-L2 image encoder and a BERT-style text encoder, emphasizing scale and weak supervision over architectural novelty.
\end{enumerate}

\section{Experiments}

\noindent
This study is structured around three key research questions. Below is a brief overview of the main findings; detailed analyses and results are provided in the subsections that follow.

\begin{enumerate}
    \item \textbf{Can LLMs generate consistent and context-aware corruption strategies?} (Section~\ref{sec:evalspec}) \\
    Yes; our approach produces interpretable and domain-relevant corruptions across diverse application settings.
    
    \item \textbf{How robust are foundation VLMs under domain-specific corruptions?} (Section~\ref{sec:evalmodels}) \\
    \clip{} is most robust overall; \alignmodel{} is weakest; \siglip{} shows domain-specific strengths but lower stability.

    \item \textbf{How does model architecture affect robustness?} (Section~\ref{sec:evalarchitecture}) \\
    Transformer models with larger capacity and QuickGELU activation show the strongest overall robustness. ViT-L/14 (QuickGELU) consistently leads across domains. Smaller patch sizes and higher input resolution can improve robustness in some cases, but effects vary by domain.

\end{enumerate}

\subsection{Can LLMs Generate Consistent and Context-Aware Corruption Strategies?}
\label{sec:evalspec}
\paragraph{Answer:} Yes, LLMs consistently produce stable and context-relevant corruption sets, supporting automated domain-aware robustness evaluation.

\begin{figure}[tbp]
    \centering
    \includegraphics[width=0.9\textwidth, alt={
    A heatmap chart showing the selection frequency of 16 image corruption types (columns) across 6 different domains (rows). The domains are Driving, Handheld, Manufacturing, Medical, People, and Satellite. Each cell contains a number from 0 to 10, indicating how often an LLM chose a specific corruption for a given domain over 10 runs. The cell background color intensity, from white to dark blue, also represents this count. 
    Certain cells are specially highlighted to show a sanity check: green boxes for "whitelisted" (plausible) corruptions and red for "blacklisted" (implausible) ones. The chart demonstrates that the LLM is context-aware, as whitelisted corruptions consistently have high counts (mostly 9 or 10), while all blacklisted corruptions have a count of 0.}]
{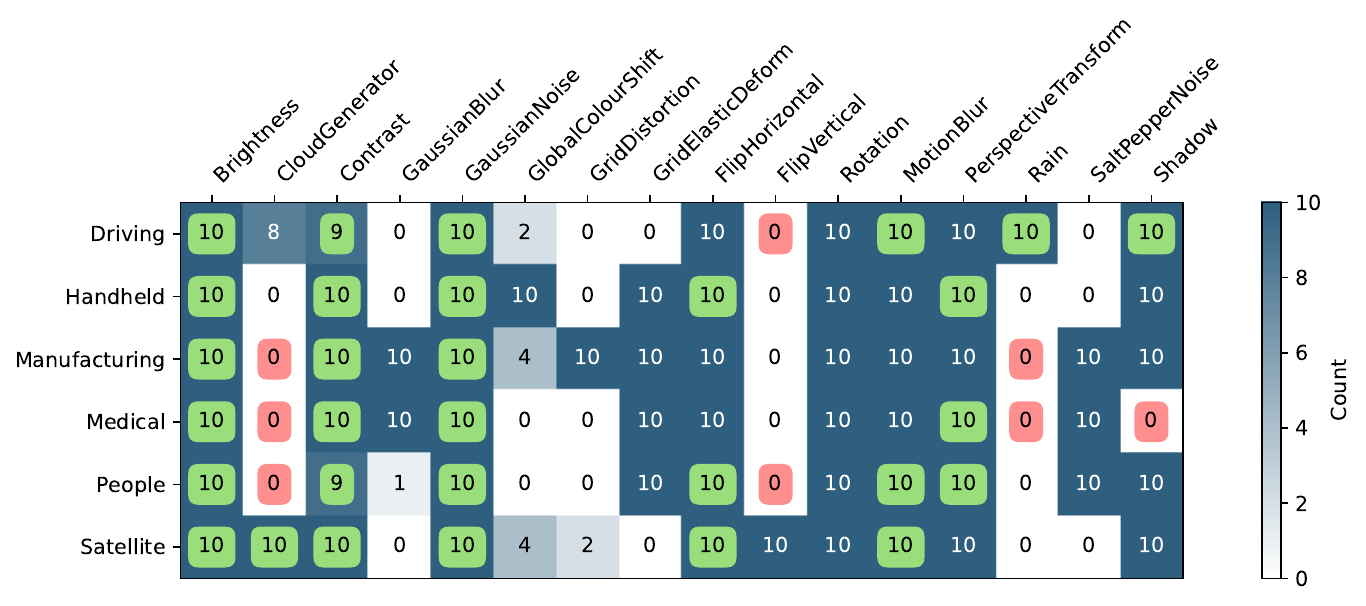}
    \caption{Heatmap of selection frequencies showing how often each corruption was chosen in 10 independent \texttt{GPT-4o} runs. As a sanity check, we defined a simple \emph{whitelist} of obviously relevant corruptions and a \emph{blacklist} of clearly implausible ones for every domain. \colorbox{customgreen}{Green} cells belong to the whitelist, whereas \colorbox{customred}{red} cells belong to the blacklist. All whitelist/blacklist rules are summarised in Table~\ref{tab:corruption_whitelist_blacklist} in \ref{app:corruption_explanations}.}

    \label{fig:corruption_selection}
\end{figure}

\paragraph{Experimental Setup:}
To assess whether large language models (LLMs) can generate stable and context-aware corruption strategies, we prompted \texttt{GPT-4o} (temperature = 0) with a domain description and the predefined list of corruptions from Tables~\ref{tab:basic_corruption_methods} and~\ref{tab:advanced_corruption_methods}. For each of the six application from \ref{par:applications}, we repeated the prompting process 10 times and recorded which corruptions were selected. This allowed us to evaluate both the stability and contextual relevance of the LLM outputs. Corruption sets for downstream robustness evaluation were determined via majority vote across these runs. 

\paragraph{Results:}
Figure~\ref{fig:corruption_selection} shows the selection frequencies of each corruption method across all domains. We observe that core transformations such as \textit{Brightness}, \textit{Contrast}, and \textit{ImageRotation} were selected consistently across domains, indicating stable LLM behavior. Additionally, the emergence of domain-specific corruptions—e.g., \textit{CloudGenerator} for Satellite imagery and \textit{GridDistortion} for Manufacturing—demonstrates that the model adapts its choices to the provided context. This confirms that LLMs can serve as a reliable component in generating tailored robustness benchmarks.

A qualitative analysis of the textual justifications generated for each corruption is provided in \ref{app:corruption_explanations}, where we assess the reasoning quality and its alignment with domain-specific visual challenges.

\subsection{How robust are foundation VLMs under domain-specific corruptions?}
\label{sec:evalmodels}

\paragraph{Answer:}
Across all evaluated domains, \clip{} consistently achieves the best mean robustness. However, in certain specialized domains \siglip{} exhibits superior relative robustness, highlighting the importance of domain-specific evaluations

\begin{figure}[tb]
    \centering
    \includegraphics[width=1.0\linewidth, alt={
A grid of ten line plots comparing the balanced accuracy of three vision-language models—CLIP (green), SigLIP (blue), and ALIGN (orange)—on a People Recognition task under various image corruptions. A red dashed line indicates the random baseline at 0.25 balanced accuracy.
Each subplot corresponds to a specific corruption type, such as Brightness, Gaussian Noise, or Rotation. The y-axis represents balanced accuracy from 0.0 to 1.0, and the x-axis represents the increasing intensity of the corruption.
A key trend shown across several plots, particularly Gaussian Noise and SaltPepperNoise, is that the ALIGN model (orange line) starts with high accuracy at low corruption levels but its performance drops sharply as corruption intensity increases. In contrast, the CLIP model (green line) generally maintains higher accuracy across increasing levels of corruption, demonstrating greater robustness.
    }]
    {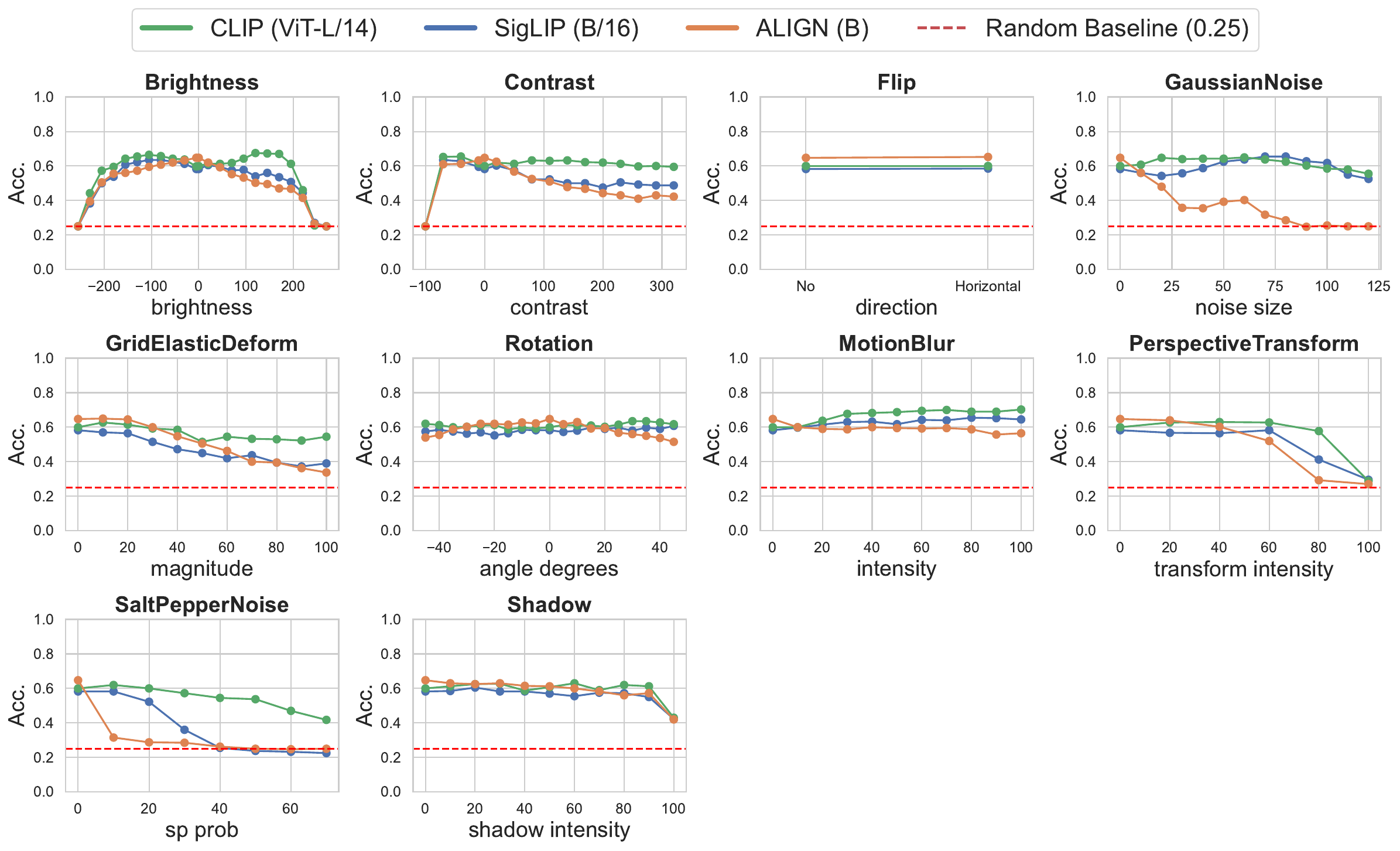}
    \caption{Balanced accuracy under various corruptions in a \textbf{People Recognition} task. \alignmodel{} leads in clean accuracy, however, drops even under moderate noise, while \clip{} remains more stable overall.}
    \label{fig:acc_people_recog}
\end{figure}

\paragraph{Experimental Setup:} 
We evaluate three large contrastive VLMs—\clip, \siglip, and \alignmodel—under a zero-shot protocol across six real-world domains (see Section~\ref{par:applications}). For each domain, a tailored corruption set is applied (Section~\ref{sec:evalspec}), encompassing geometric, photometric, and environment-specific disturbances. We test multiple corruption intensities and primarily measure \emph{balanced accuracy}. In addition, we calculate aggregate robustness metrics including mCE, rCE, and mFR as defined in Section~\ref{par:metrics}. A summary of all results is provided in Table~\ref{tab:foundation_vlm_metrics} (see \ref{app:foundation_vlm_metrics}), and Figure~\ref{fig:acc_people_recog} presents representative results for the \textbf{People Recognition} domain. Detailed per-domain plots for all six scenarios can be found in \ref{app:robustness_plots}.

\paragraph{Results:}
\clip{} consistently achieves the lowest mCE across all domains, indicating strong overall robustness. 
However, some domains exhibit nuanced trends: in \textbf{Medical Diagnosis} and \textbf{Manufacturing Quality}, \siglip{} surpasses \clip{} in rCE. 
Notably, \alignmodel{}, despite achieving high clean accuracy, experiences severe performance degradation under mild noise corruptions (rCE up to 14.78). 
Figure~\ref{fig:acc_people_recog} highlights this effect in \textbf{People Recognition}, where \alignmodel{}'s accuracy sharply declines with minimal noise, approaching random-chance levels, whereas \clip{} and \siglip{} remain robust. Additionally, brightness and contrast changes only slightly affect \clip{}, but significantly degrade \alignmodel{} performance beyond moderate distortion levels.


\subsection{How Does Model Architecture Affect Robustness?}
\label{sec:evalarchitecture}

\paragraph{Answer:}
Transformer-based architectures generally offer superior robustness compared to ResNet baselines. ViT-L/14 (QuickGELU) consistently ranks best or near-best across domains in terms of robustness. While smaller patch sizes can improve robustness in some cases, this effect is not consistent across all domains. Similarly, higher input resolution leads to lower robustness on every domain.

\begin{table}[tbp]
\centering
\scriptsize 
\caption{Per-domain clean balanced accuracy (Acc.) and mCE for architectural variants of \clip{}  relative to \texttt{CLIP (ViT-L/14)}}
\label{tab:architecture_summary}
\begin{tabular}{p{0.22\textwidth}>{\centering\arraybackslash}p{0.09\textwidth}>{\centering\arraybackslash}p{0.09\textwidth}>{\centering\arraybackslash}p{0.09\textwidth}>{\centering\arraybackslash}p{0.09\textwidth}>{\centering\arraybackslash}p{0.09\textwidth}>{\centering\arraybackslash}p{0.09\textwidth}}
\toprule
 & \multicolumn{2}{c}{\textbf{Driving}} & \multicolumn{2}{c}{\textbf{Handheld}} & \multicolumn{2}{c}{\textbf{Manufacturing}} \\
 & Acc.\,$\uparrow$ & mCE\,$\downarrow$ & Acc.\,$\uparrow$ & mCE\,$\downarrow$ & Acc.\,$\uparrow$ & mCE\,$\downarrow$ \\
\midrule
ResNet101 & 0.66 & 1.27 & 0.72 & 1.48 & 0.77 & 0.83 \\
ResNet50 & 0.66 & 1.25 & 0.73 & 1.52 & 0.65 & 1.05 \\
ViT-B/16 & 0.74 & 1.01 & 0.81 & 1.25 & 0.65 & 0.98 \\
ViT-B/32 & \textbf{0.82} & \textbf{0.84} & 0.71 & 1.51 & 0.61 & 1.10 \\
ViT-L/14 (QGelu) & 0.74 & 0.87 & 0.86 & \textbf{0.88} & \textbf{0.78} & \textbf{0.74} \\
ViT-L/14 (baseline) & 0.69 & 1.00 & 0.85 & 1.00 & 0.62 & 1.00 \\
ViT-L/14-336 & 0.74 & 0.85 & \textbf{0.87} & 0.90 & 0.70 & 0.91 \\
\midrule
 & \multicolumn{2}{c}{\textbf{Medical}} & \multicolumn{2}{c}{\textbf{People}} & \multicolumn{2}{c}{\textbf{Satellite}} \\
 & Acc.\,$\uparrow$ & mCE\,$\downarrow$ & Acc.\,$\uparrow$ & mCE\,$\downarrow$ & Acc.\,$\uparrow$ & mCE\,$\downarrow$ \\
\midrule
ResNet101 & 0.63 & 1.61 & 0.62 & 1.35 & 0.75 & 1.49 \\
ResNet50 & 0.66 & 1.64 & 0.68 & 1.28 & 0.70 & 1.65 \\
ViT-B/16 & 0.78 & 1.38 & 0.66 & 1.22 & 0.80 & 1.23 \\
ViT-B/32 & 0.81 & 1.32 & \textbf{0.70} & 1.11 & 0.75 & 1.40 \\
ViT-L/14 (QGelu) & \textbf{0.97} & \textbf{0.70} & 0.60 & 1.20 & 0.85 & \textbf{0.84} \\
ViT-L/14 (baseline) & 0.85 & 1.00 & \textbf{0.70} & 1.00 & 0.84 & 1.00 \\
ViT-L/14-336 & 0.86 & 0.91 & 0.68 & \textbf{0.97} & \textbf{0.87} & 0.92 \\
\bottomrule
\end{tabular}
\end{table}

\paragraph{Experimental Setup:}
We compare a set of \clip{} model variants with architectural differences in backbone type (ResNet vs. Vision Transformer), patch size, input resolution, and activation function. All models were trained by OpenAI on the same dataset, ensuring consistent pretraining conditions. 

Each model is evaluated in a zero-shot setting across six real-world application domains (Section~\ref{par:applications}), using the tailored domain-specific corruption sets from Section~\ref{sec:evalspec}. For each corruption type and intensity, we compute balanced accuracy. We also report aggregate robustness metrics: mCE, rCE, and mFR, as defined in Section~\ref{par:metrics}. A complete summary is provided in Table~\ref{tab:foundation_vlm_metrics} (\ref{app:architecture_models_metrics}), with selected results presented in Table~\ref{tab:architecture_summary}.

\paragraph{Results:}
Table~\ref{tab:architecture_summary} shows clean accuracy and mCE for each architecture across all six domains. Several consistent patterns emerge:

\begin{itemize}
    \item \textbf{Transformer vs. ResNet:} Vision Transformer (ViT) models substantially outperform ResNet-based architectures in both clean accuracy and robustness. ResNet-50 and ResNet-101 show the lowest accuracy and highest average mCE across domains.

    \item \textbf{Patch size:} ViT-B/16 does not consistently outperform ViT-B/32. In both clean accuracy and mCE, there is no clear advantage for one of the two architectures. 

    \item \textbf{Resolution:} ViT-L/14@336 achieves lower mCE than ViT-L/14 (baseline) in all domains, indicating improved robustness with higher input resolution. Gains in clean accuracy are observed in 5 of the 6 application scenarios.
    
    \item \textbf{Activation function:} The ViT-L/14 variant with QGelu activation achieves  better clean accuracy and mCE then the ViT-L/14 baseline in 5 out of 6 cases. 

    \item \textbf{Best overall model:} ViT-L/14 (QGelu) emerges as the most robust architecture in our comparison, achieving the lowest mCE in 4 and highest clean accuracy in 2 application scenarios.
\end{itemize}

\noindent
Full results including additional robustness metrics are provided in \ref{app:architecture_models_metrics}.

\section{Conclusion and Future Work}
\label{sec:conclusion}
In this work, we introduced \deepbench{}, a modular, label-free framework for evaluating the domain-specific robustness of VLMs. Unlike traditional robustness benchmarks that focus on synthetic or generic corruptions, \deepbench{} employs large language models to generate realistic, scenario-specific corruptions, enabling scalable, zero-shot robustness assessments without requiring labeled data.

Our experiments across six real-world domains highlight several insights. Among contrastive VLMs, \clip{} consistently achieves superior robustness compared to \siglip{} and \alignmodel{}. Architectural choices also significantly influence robustness: Transformer models with finer patches, higher resolution, and QuickGELU activation generally outperform other variants. These findings emphasize the importance of targeted evaluations beyond conventional benchmarks.

\paragraph{Limitations.} \deepbench{} currently uses a fixed set of corruption primitives and relies on LLM-generated context interpretations. Our analysis focuses primarily on image classification for contrastive models and does not yet fully address autoregressive VLMs, whose robustness heavily depends on prompt design and output parsing.

\paragraph{Future Work.} Future work will extend \deepbench{} beyond classification tasks, incorporate adaptive corruption sampling, and include temporal or multi-frame corruptions suitable for video and robotics scenarios. We also plan to systematically evaluate autoregressive models (e.g., \llava), investigating the impact of prompt strategies and output handling on their robustness.

\begin{credits}
\subsubsection{\ackname}
This research was funded by the \textit{Deutsche Forschungsgemeinschaft (DFG, German Research Foundation)} via the Project ApplFM (\texttt{528483508}) and the \textit{Institut für angewandte Forschung Berlin (IFAF, Berlin Institute for Applied Research)} via Project TrustAdHocAI (TAHAI).

\subsubsection{\discintname}
The authors declare no competing financial or non-financial interests related to this work. This research was conducted independently and is not affiliated with or endorsed by any of the organizations that developed the evaluated models or datasets.

\end{credits}

\appendix
\renewcommand{\thesection}{Appendix \Alph{section}}

\section{Prompts}
\label{app:prompts}

To guide the LLM in selecting appropriate image corruptions for a given application domain, we use a structured text prompt that includes:
\begin{itemize}
    \item A brief role assignment (e.g., "You are an expert in data augmentation for deep learning.")
    \item A description of the domain, including image source and task context
    \item A list of available augmentation operations with brief descriptions
    \item Explicit instructions on the expected output format
\end{itemize}

An example prompt for the medical imaging domain is shown in Listing~\ref{lst:prompting_example}.

\begin{lstlisting}[
  language=Clean,
  caption={Prompting template for domain-specific corruption selection},
  label=lst:prompting_example,
  basicstyle=\tiny\ttfamily,
  breaklines=true,
  breakatwhitespace=false,
  columns=fullflexible,
  keepspaces=true,
  showstringspaces=false,
  frame=single,
  xleftmargin=1em,
  xrightmargin=1em,
  aboveskip=1em,
  belowskip=1em
]
[User] 
You are an expert in data augmentation for deep learning. 

I need augmentation recommendations for the following domain:
"The knee osteoarthritis dataset is sourced from the Osteoarthritis Initiative (OAI) and contains frontal X-ray images of knee joints, showing the bone structure and joint space, which are labeled according to the Kellgren-Lawrence (KL) grading system."

Provide a list of augmentation names with a brief explanation of why each is useful in this domain.

The following augmentations are available:
- Shadow: Adds synthetic shadows to an image to simulate lighting conditions.
- PerspectiveTransformation: Warps the image by changing its perspective, as if viewed from a different angle.
- GridDistortion: Distorts the image by applying a grid-like warping effect, bending specific areas.
- ImageFlipHorizontal: Flips the image along the vertical axis, mirroring it horizontally.
- ImageFlipVertical: Flips the image along the horizontal axis, mirroring it vertically.
- SaltPepperNoise: Adds random white and black dots to the image, mimicking noisy pixels.
- Contrast: Alters the difference between light and dark areas to make the image appear more or less vivid.
- Brightness: Changes the overall lightness or darkness of the image.
- ImageRotation: Rotates the image by a specified angle, keeping its contents intact.
- GaussianNoise: Adds random, fine-grained noise following a Gaussian distribution to simulate sensor noise.
- GridElasticDeformation: Applies a rubber-sheet-like deformation to the image, bending it smoothly.
- MotionBlur: Blurs the image to simulate movement, as if the camera or object was in motion.
- GaussianBlur: Smoothens the image by blurring it, reducing fine details or noise.
- GlobalColourShift: Adjusts the overall color balance of the image, shifting its tones globally.
- Rain: Adds synthetic raindrop effects or streaks to mimic rainy conditions.
- CloudGenerator: Overlays or generates cloud-like textures in the image, simulating an overcast sky.

Using these augmentations, provide recommendations in the following format:

Example:
1. HistEqualization: Enhancing contrast through histogram equalization is crucial for highlighting subtle differences in knee joint structures. This helps the model identify small variations indicative of arthritis progression.
2. GaussianBlur: Simulates blurring effects that may occur in real-world imaging, helping the model learn to handle reduced detail while still identifying key features.
3. ImageRotation: Slightly rotating images introduces variability to account for different imaging angles. This improves the model's robustness to positional variations in medical scans.

Now, generate recommendations specific to the domain provided.
\end{lstlisting}

\section{LLM Explanations for Domain-Specific Corruption Selection}
\label{app:corruption_explanations}

To automatically generate corruption strategies tailored to each application domain, we prompt \texttt{GPT-4o} with structured templates (see \ref{app:prompts}). Each prompt includes (i) a domain description, (ii) a list of candidate corruptions with semantic descriptions (from Tables~\ref{tab:basic_corruption_methods} and~\ref{tab:advanced_corruption_methods}), and (iii) instructions for selecting relevant corruptions and explaining their relevance.

\begin{table}
\centering
\caption{Corruptions that should always or never be selected for specific domains. This curated whitelist/blacklist reflects intuitive domain knowledge}
\label{tab:corruption_whitelist_blacklist}
{
\scriptsize\ttfamily
\begin{tabular}{p{0.28\textwidth}p{0.33\textwidth}p{0.33\textwidth}}
\toprule
\textbf{Corruption Type} & \textbf{Whitelist:} & \textbf{Blacklist:} \\
\midrule
\textbf{Brightness} & All domains & — \\ \midrule
\textbf{Cloud Overlay} & Satellite & Medical, Manufacturing, People \\ \midrule
\textbf{Contrast} & All domains & — \\ \midrule
\textbf{Gaussian Noise} & All domains & — \\ \midrule
\textbf{Flip (Horizontal)} & People, Satellite, Handheld &  \\  \midrule
\textbf{Flip (Vertical)} &  & Driving, People \\  \midrule
\textbf{Motion Blur} & Driving, Satellite, People & \\ \midrule
\textbf{Perspective Transform.} & Medical, People, Handheld &  \\ \midrule
\textbf{Rain} & Driving & Medical, Manufacturing \\ \midrule
\textbf{Shadow} & Driving & Medical \\ 
\bottomrule
\end{tabular}
}
\end{table}

The model's responses consist of configuration blocks listing selected augmentations, each accompanied by a short explanation of its contextual relevance. To assess whether the LLM selects perturbations in a semantically meaningful way, we evaluate the resulting corruption sets along two axes:

\begin{itemize}
    \item \textbf{Selection accuracy:} Does the LLM consistently choose perturbations that are clearly relevant for the domain?
    \item \textbf{Exclusion accuracy:} Does it avoid obviously implausible or irrelevant augmentations for the setting?
\end{itemize}

Figure~\ref{fig:corruption_selection} shows a heatmap of selected augmentations across six domains. The results are consistent and interpretable: brightness shifts, for example, are selected across nearly all domains due to their universal relevance, while implausible corruptions such as rain in medical imaging or cloud overlay in manufacturing are never chosen.

We summarize these expectations as a domain-informed \textbf{whitelist/blacklist} in Table~\ref{tab:corruption_whitelist_blacklist}, showing corruption-domain pairings that should always or never be selected. These rules are fully satisfied by the LLM’s output, indicating strong semantic alignment.

In addition to selecting plausible corruptions, the LLM also produces coherent justifications for its choices. Table~\ref{tab:llm_explanations} presents a sample of these explanations, demonstrating domain awareness and the ability to ground augmentation choices in real-world deployment conditions.

\begin{table}
\centering
\scriptsize
\caption{Example explanations provided by the LLM (\texttt{GPT-4o}) for selecting specific corruptions in different domains. These illustrate the contextual relevance and semantic grounding of the corruption strategies.}
\label{tab:llm_explanations}
\begin{tabular}{p{0.9\linewidth}}
\toprule
\textbf{Satellite Imaging:} \\
\textit{“Adding synthetic shadows can simulate different lighting conditions that occur naturally in aerial images. This helps the model become robust to variations in sunlight and shadow patterns, which are common in outdoor scenes.”} \\
\midrule
\textbf{People Recognition:} \\
\textit{“Flipping images horizontally is a simple yet effective way to increase dataset diversity. Since facial expressions are generally symmetrical, this augmentation helps the model learn to recognize expressions regardless of the face's orientation.”} \\
\midrule
\textbf{Medical Diagnosis:} \\
\textit{“Adding salt and pepper noise can simulate pixel-level noise that may occur in digital imaging. This helps the model learn to identify important features even in the presence of such noise, improving its robustness to image artifacts.”} \\
\midrule
\textbf{Autonomous Driving:} \\
\textit{“Adding synthetic shadows can simulate various lighting conditions encountered in real-world driving scenarios. This helps the model become robust to changes in illumination, which is crucial for recognizing objects under different lighting conditions.”} \\
\midrule
\textbf{Handheld:} \\
\textit{“Simulating motion blur helps the model learn to recognize food items even when there is slight movement during photography, such as a shaky hand or moving camera. This improves the model's robustness to motion-related artifacts.”} \\
\bottomrule
\end{tabular}
\end{table}

\section{Additional Robustness Plots}
\label{app:robustness_plots}

This appendix provides additional robustness evaluation plots that complement Section~\ref{sec:evalmodels}. Each figure visualizes how balanced accuracy degrades under increasing corruption severity for one of the six application domains discussed in our main experiments. These visualizations support the domain-specific comparisons referenced in the quantitative results (Table~\ref{tab:foundation_vlm_metrics}) and the per-domain robustness analysis in Section~\ref{sec:evalmodels}.

\begin{figure}
    \centering
    \includegraphics[width=\textwidth, alt={
A grid of ten line plots comparing the balanced accuracy of three models—CLIP (green), SigLIP (blue), and ALIGN (orange)—in an autonomous driving domain under various corruptions. Each plot shows balanced accuracy (y-axis) versus the intensity of a specific corruption (x-axis). A red dashed line at 0.25 balanced accuracy represents the random baseline.
The plots show that while all models perform similarly for corruptions like Contrast and Rotation, their robustness varies significantly for others. Notably, in the plots for gaussian noise, perspective transform, and rain, the CLIP model (green line) maintains the highest balanced accuracy as corruption intensity increases. The performance of SigLIP (blue line) and ALIGN (orange line) degrades much more rapidly under these same conditions, indicating lower robustness in scenarios critical for autonomous driving.
    }]
    {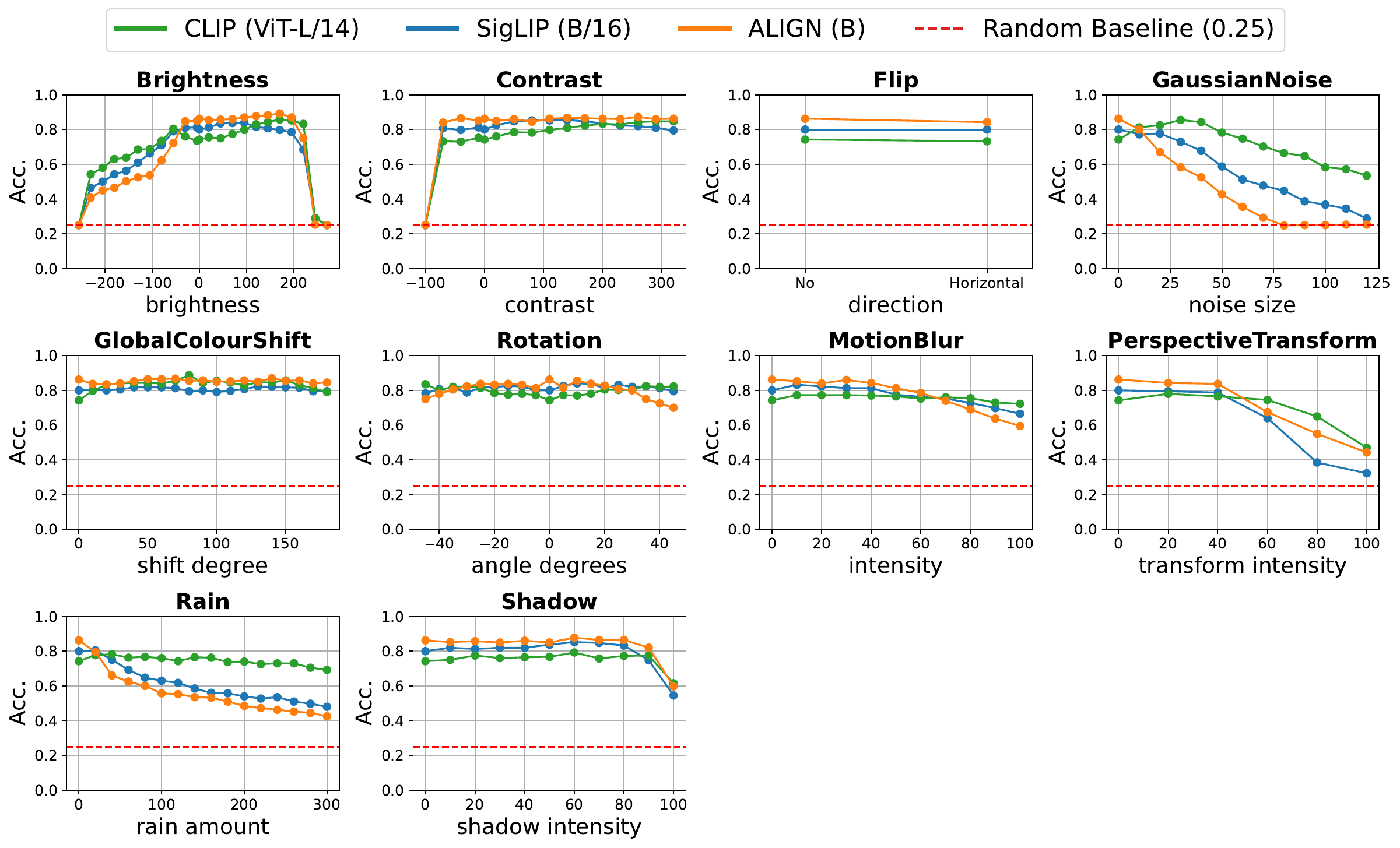}
    \caption{Relative balanced accuracy under corruption for the \textbf{Autonomous Driving} domain.}
    \label{fig:robustness_autonomous}
\end{figure}

\begin{figure}
    \centering
    \includegraphics[width=\textwidth, alt={
A grid of twelve line plots comparing the balanced accuracy of three models—CLIP (green), SigLIP (blue), and ALIGN (orange)—on a manufacturing quality control task. Each subplot shows balanced accuracy versus the intensity of one of twelve different image corruptions, such as gaussian blur, perspective transformation, and salt and pepper noise. A red dashed line indicates the random baseline at 0.07 balanced accuracy.
The plots consistently show that CLIP is the most robust model, maintaining higher balanced accuracy across most types and intensities of corruption, particularly for rotation and blurs. ALIGN is generally the least robust, with its balanced accuracy degrading more quickly than the others. The salt and pepper noise plot is a particularly stark example: all models' accuracies drop, but ALIGN's drops almost immediately to the random baseline, highlighting a significant vulnerability.
    }]
    {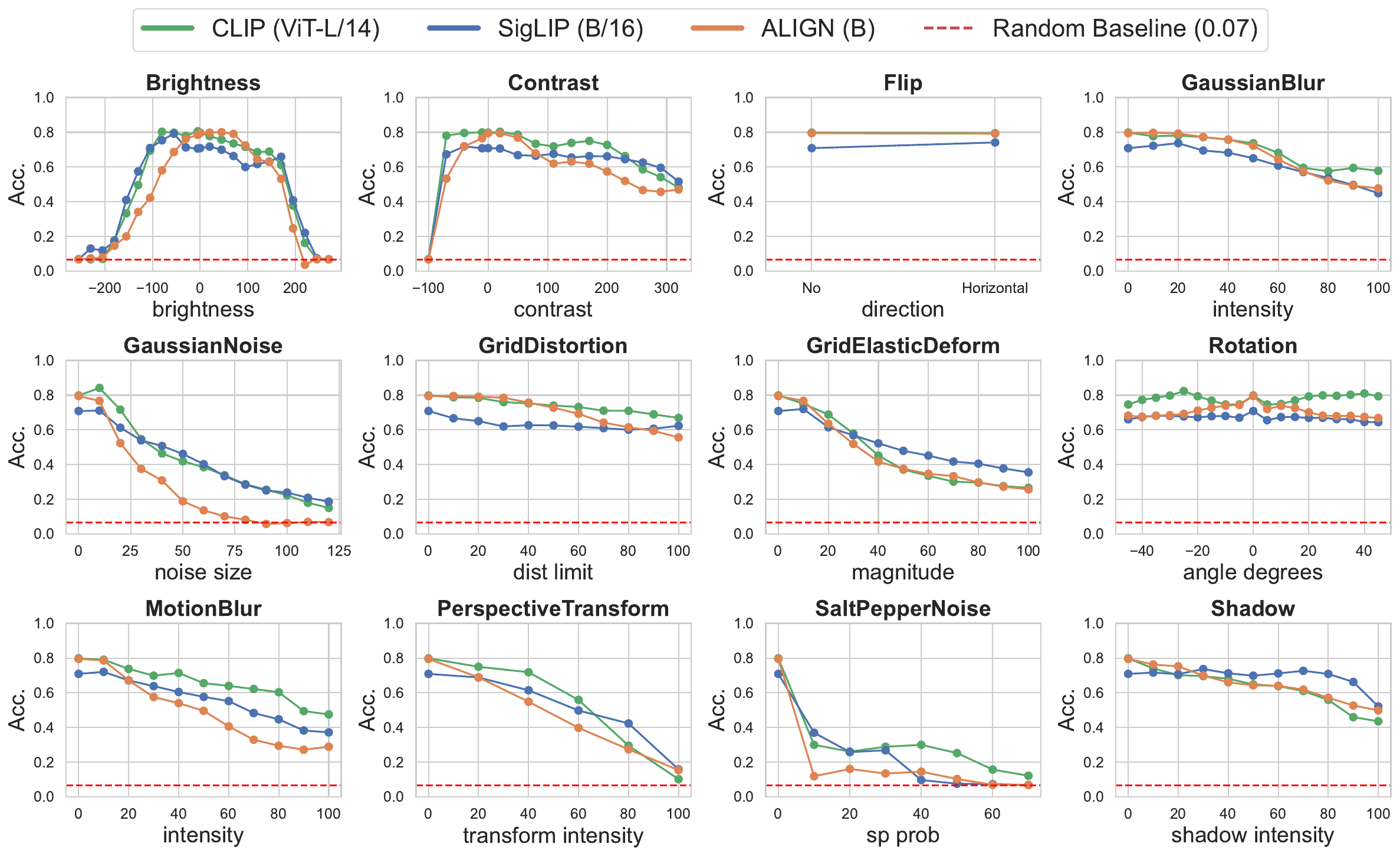}
    \caption{Relative balanced accuracy under corruption for the \textbf{Manufacturing Quality} domain}
    \label{fig:robustness_manufacturing}
\end{figure}

\begin{figure}
    \centering
    \includegraphics[width=\textwidth, alt={
A grid of ten line plots comparing the balanced accuracy of three models—CLIP (green), SigLIP (blue), and ALIGN (orange)—on a medical diagnosis task. Each subplot displays balanced accuracy versus the intensity of a specific corruption, such as brightness or gaussian noise. A red dashed line shows the random baseline at 0.12 balanced accuracy.
The plots consistently show that the CLIP model is the most robust, with its green line maintaining the highest balanced accuracy across nearly all corruption types and intensities. The performance gap is especially notable in the brightness, contrast, and rotation plots. While all models' accuracies degrade with increasing noise or distortion, CLIP's performance degrades the slowest, indicating its superior stability for the medical diagnosis domain.
    }]
    {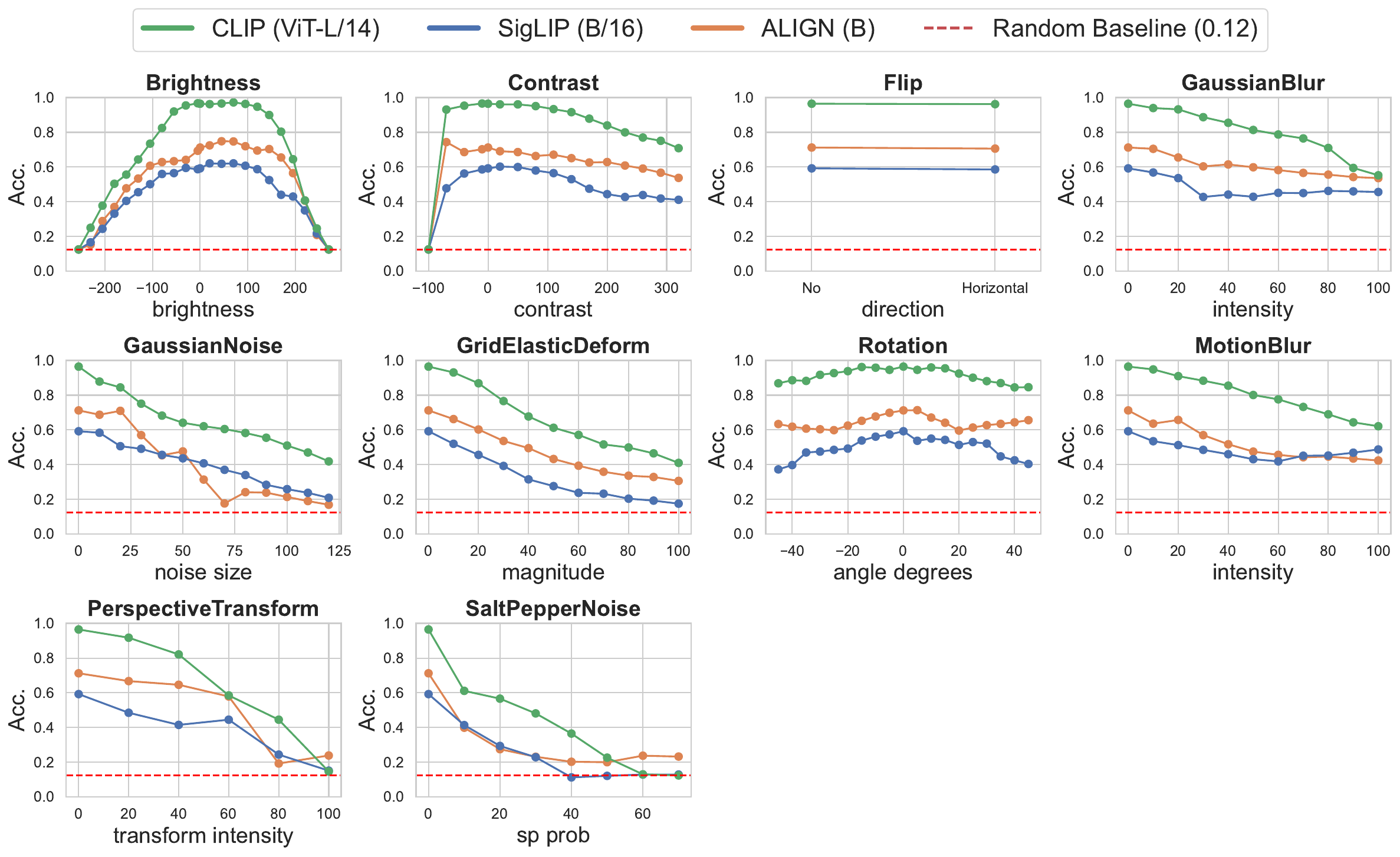}
    \caption{Relative balanced accuracy under corruption for the \textbf{Medical Diagnosis} domain}
    \label{fig:robustness_medical}
\end{figure}

\begin{figure}
    \centering
    \includegraphics[width=\textwidth, alt={
A grid of nine line plots comparing the balanced accuracy of three models (CLIP in green, SigLIP in blue, ALIGN in orange) on a Satellite Imaging task. Each subplot shows balanced accuracy on the y-axis against the intensity of a specific corruption on the x-axis, such as cloud generator, rotation, or shadow. A red dashed line indicates the random baseline at 0.07 balanced accuracy.
The figure clearly demonstrates that the CLIP model is significantly more robust than the others in this domain. Its green line is consistently positioned above the blue and orange lines across nearly all plots. This performance gap is especially pronounced for domain-specific corruptions like cloud generator and general transformations like rotation, where CLIP's balanced accuracy remains high and stable while the other models' performance degrades substantially.
    }]
    {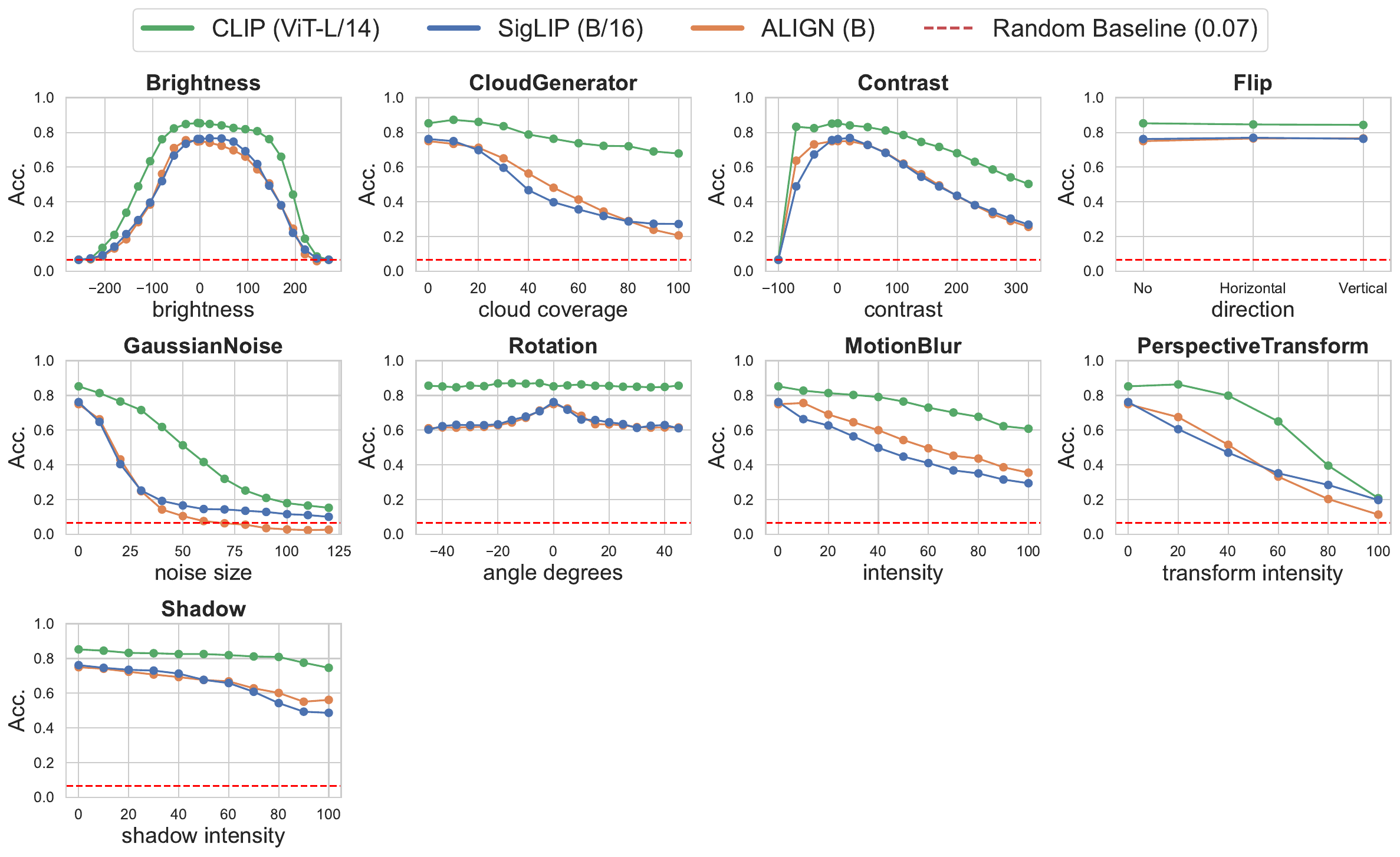}
    \caption{Relative balanced accuracy under corruption for the \textbf{Satellite Imaging} domain}
    \label{fig:robustness_satellite}
\end{figure}

\begin{figure}
    \centering
    \includegraphics[width=\textwidth, alt={
A grid of ten line plots comparing the balanced accuracy of three models—CLIP (green), SigLIP (blue), and ALIGN (orange)—on a Handheld domain task. Each subplot shows balanced accuracy versus the intensity of a specific corruption. A red dashed line at 0.01 balanced accuracy represents the random baseline.
Across nearly all corruption types, the CLIP model consistently demonstrates superior robustness, with its green line positioned highest. SigLIP generally shows the lowest performance. The performance difference is particularly severe for corruptions like gaussian noise and perspective transformation, where CLIP's balanced accuracy degrades slowly, while the accuracies of SigLIP and ALIGN fall rapidly toward the random baseline.
    }]
    {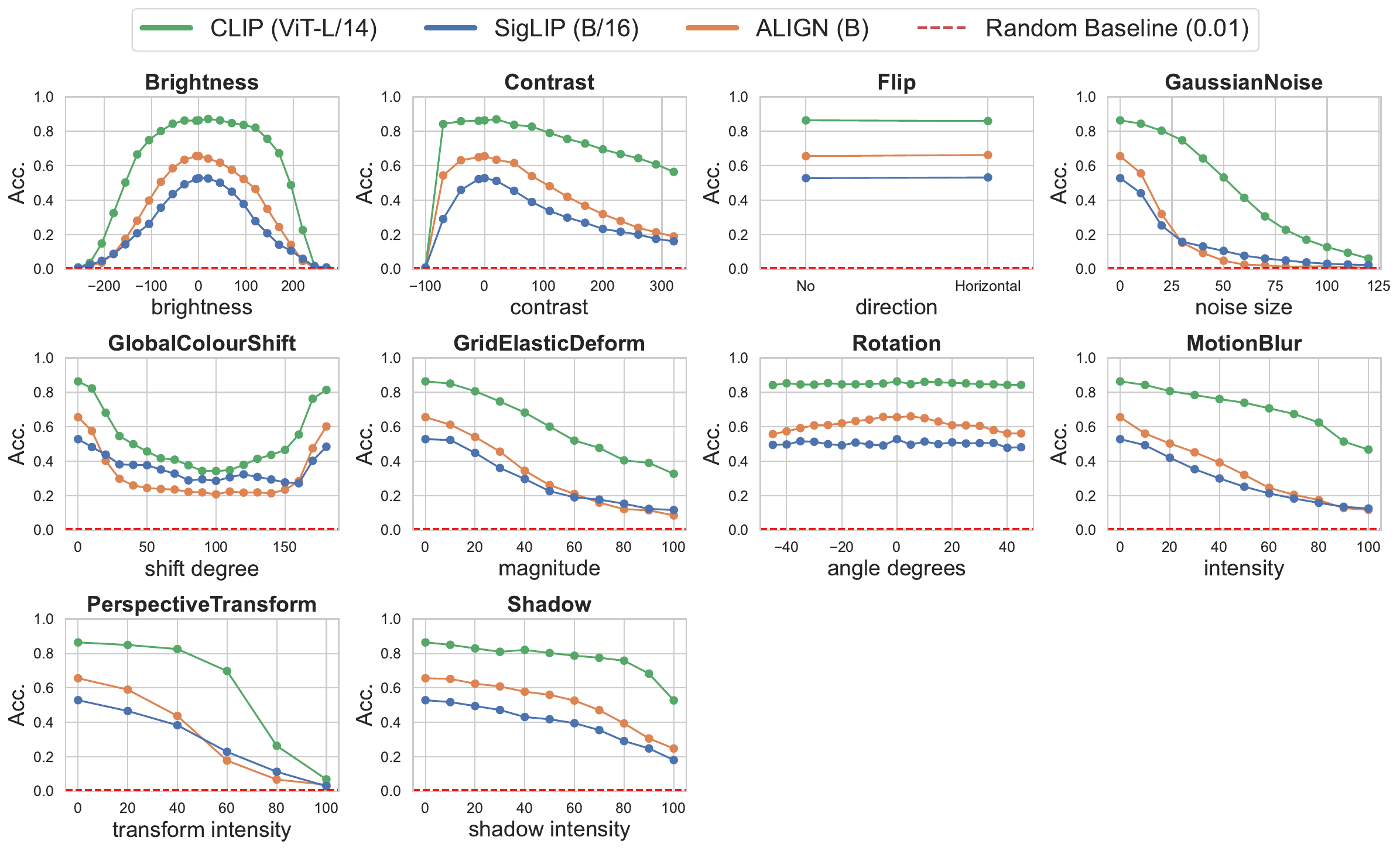}
    \caption{Relative balanced accuracy under corruption for the \textbf{Handheld} domain}
    \label{fig:robustness_handheld}
\end{figure}

\section{Detailed Results for Foundation Vision-Language Models}
\label{app:foundation_vlm_metrics}

Table~\ref{tab:foundation_vlm_metrics} provides the full quantitative results for the robustness evaluation of three foundation VLMs (Section~\ref{sec:evalmodels}). For each domain, we report the clean accuracy and relative robustness metrics (mCE, rCE, and mFR), all computed with respect to the baseline model \texttt{CLIP (ViT-L/14)}.

\begin{table}
\centering
\tiny 
\caption{Robustness evaluation of foundation VLMs across application domains with clean balanced accuracy (Acc.) and mCE, rCE and mFR relative to \texttt{CLIP (ViT-L/14)}. \( r \) is the Pearson correlation between balanced accuracy scores and label flip probability.}
\label{tab:foundation_vlm_metrics}
\begin{tabular}{p{0.25\textwidth}>{\centering\arraybackslash}p{0.07\textwidth}>{\centering\arraybackslash}p{0.07\textwidth}>{\centering\arraybackslash}p{0.07\textwidth}>{\centering\arraybackslash}p{0.07\textwidth}>{\centering\arraybackslash}p{0.07\textwidth}}
\toprule
 & Acc.\,$\uparrow$ & mCE\,$\downarrow$ & rCE\,$\downarrow$ & mFR\,$\downarrow$ & $r$\,$\downarrow$ \\
\midrule
\multicolumn{6}{l}{\textbf{AutonomousDriving}} \\
ALIGN (B) & \textbf{0.86} & 1.05 & 10.39 & 0.91 & \textbf{-0.99} \\
CLIP (ViT-L/14) & 0.74 & \textbf{1.00} & \textbf{1.00} & 1.00 & -0.79 \\
SigLip (B/16) & 0.80 & 1.09 & 5.94 & \textbf{0.88} & -0.96 \\
\midrule
\multicolumn{6}{l}{\textbf{Handheld}} \\
ALIGN (B) & 0.65 & 1.92 & 1.59 & 2.03 & -0.98 \\
CLIP (ViT-L/14) & \textbf{0.86} & \textbf{1.00} & \textbf{1.00} & \textbf{1.00} & \textbf{-1.00} \\
SigLip (B/16) & 0.53 & 2.21 & 1.20 & 1.84 & -0.97 \\
\midrule
\multicolumn{6}{l}{\textbf{ManufacturingQuality}} \\
ALIGN (B) & \textbf{0.80} & 1.16 & 1.56 & 1.09 & \textbf{-0.99} \\
CLIP (ViT-L/14) & \textbf{0.80} & \textbf{1.00} & 1.00 & \textbf{1.00} & -0.98 \\
SigLip (B/16) & 0.71 & 1.16 & \textbf{0.82} & 1.22 & -0.93 \\
\midrule
\multicolumn{6}{l}{\textbf{MedicalDiagnosis}} \\
ALIGN (B) & 0.71 & 2.64 & 1.03 & 1.73 & -0.95 \\
CLIP (ViT-L/14) & \textbf{0.97} & \textbf{1.00} & 1.00 & \textbf{1.00} & \textbf{-1.00} \\
SigLip (B/16) & 0.59 & 3.42 & \textbf{0.99} & 2.47 & -0.93 \\
\midrule
\multicolumn{6}{l}{\textbf{PeopleRecognition}} \\
ALIGN (B) & \textbf{0.65} & 1.20 & 14.78 & 1.13 & \textbf{-0.94} \\
CLIP (ViT-L/14) & 0.60 & \textbf{1.00} & \textbf{1.00} & \textbf{1.00} & -0.80 \\
SigLip (B/16) & 0.58 & 1.14 & 2.58 & 1.27 & -0.82 \\
\midrule
\multicolumn{6}{l}{\textbf{SatelliteImaging}} \\
ALIGN (B) & 0.75 & 1.75 & 13.27 & 1.58 & \textbf{-0.99} \\
CLIP (ViT-L/14) & \textbf{0.85} & \textbf{1.00} & \textbf{1.00} & \textbf{1.00} & \textbf{-0.99} \\
SigLip (B/16) & 0.76 & 1.77 & 14.24 & 1.81 & -0.98 \\
\bottomrule
\end{tabular}
\end{table}

\section{Detailed Results on Model Architecture}
\label{app:architecture_models_metrics}

\noindent
Table~\ref{tab:architecture_models_metrics} reports the full robustness results for \clip{} variants with different architectural configurations, as discussed in Section~\ref{sec:evalarchitecture}. All models are trained on the same dataset by \texttt{OpenAI}, ensuring consistent pretraining conditions, but differ in backbone type (ResNet vs. ViT), input resolution, patch size, and activation function.
We present clean accuracy, mCE, rCE, and mFR across six application scenarios. These extended results complement the summary in the main paper.

\begin{table}
\centering
\tiny 
\caption{Robustness evaluation across \clip{} architectural variants with clean balanced accuracy (Acc.) and mCE, rCE and mFR relative to \texttt{CLIP (ViT-L/14)}. \( r \) is the Pearson correlation between balanced accuracy scores and label flip probability.}
\label{tab:architecture_models_metrics}
\begin{tabular}{p{0.25\textwidth}>{\centering\arraybackslash}p{0.07\textwidth}>{\centering\arraybackslash}p{0.07\textwidth}>{\centering\arraybackslash}p{0.07\textwidth}>{\centering\arraybackslash}p{0.07\textwidth}>{\centering\arraybackslash}p{0.07\textwidth}}
\toprule
 & Acc.\,$\uparrow$ & mCE\,$\downarrow$ & rCE\,$\downarrow$ & mFR\,$\downarrow$ & $r$\,$\downarrow$ \\
\midrule
\multicolumn{6}{l}{\textbf{AutonomousDriving}} \\
ResNet101 & 0.66 & 1.27 & 3.74 & 0.97 & -0.85 \\
ResNet50 & 0.66 & 1.25 & 2.37 & 1.16 & -0.82 \\
ViT-B/16 & 0.74 & 1.01 & 2.38 & \textbf{0.75} & -0.91 \\
ViT-B/32 & \textbf{0.82} & \textbf{0.84} & 6.59 & \textbf{0.75} & \textbf{-0.96} \\
ViT-L/14 (QGelu) & 0.74 & 0.87 & 1.06 & 0.98 & -0.80 \\
ViT-L/14 (baseline) & 0.69 & 1.00 & \textbf{1.00} & 1.00 & -0.75 \\
ViT-L/14-336 & 0.74 & 0.85 & 1.64 & 0.82 & -0.81 \\
\midrule
\multicolumn{6}{l}{\textbf{Handheld}} \\
ResNet101 & 0.72 & 1.48 & 1.19 & 1.41 & -0.99 \\
ResNet50 & 0.73 & 1.52 & 1.32 & 1.47 & -0.99 \\
ViT-B/16 & 0.81 & 1.25 & 1.31 & 1.23 & \textbf{-1.00} \\
ViT-B/32 & 0.71 & 1.51 & 1.25 & 1.45 & -0.99 \\
ViT-L/14 (QGelu) & 0.86 & \textbf{0.88} & \textbf{0.83} & 0.86 & \textbf{-1.00} \\
ViT-L/14 (baseline) & 0.85 & 1.00 & 1.00 & 1.00 & \textbf{-1.00} \\
ViT-L/14-336 & \textbf{0.87} & 0.90 & 0.93 & \textbf{0.85} & \textbf{-1.00} \\
\midrule
\multicolumn{6}{l}{\textbf{ManufacturingQuality}} \\
ResNet101 & 0.77 & 0.83 & 1.59 & 1.05 & \textbf{-0.99} \\
ResNet50 & 0.65 & 1.05 & 1.70 & 1.27 & -0.97 \\
ViT-B/16 & 0.65 & 0.98 & 1.12 & 0.95 & -0.97 \\
ViT-B/32 & 0.61 & 1.10 & 2.05 & 1.37 & -0.96 \\
ViT-L/14 (QGelu) & \textbf{0.78} & \textbf{0.74} & 1.20 & \textbf{0.85} & -0.98 \\
ViT-L/14 (baseline) & 0.62 & 1.00 & \textbf{1.00} & 1.00 & -0.95 \\
ViT-L/14-336 & 0.70 & 0.91 & 1.47 & 0.93 & -0.98 \\
\midrule
\multicolumn{6}{l}{\textbf{MedicalDiagnosis}} \\
ResNet101 & 0.63 & 1.61 & \textbf{0.62} & 1.76 & -0.88 \\
ResNet50 & 0.66 & 1.64 & 1.19 & 1.57 & -0.88 \\
ViT-B/16 & 0.78 & 1.38 & 2.09 & 1.40 & -0.99 \\
ViT-B/32 & 0.81 & 1.32 & 3.15 & 1.53 & -0.99 \\
ViT-L/14 (QGelu) & \textbf{0.97} & \textbf{0.70} & 1.76 & \textbf{0.81} & \textbf{-1.00} \\
ViT-L/14 (baseline) & 0.85 & 1.00 & 1.00 & 1.00 & -0.97 \\
ViT-L/14-336 & 0.86 & 0.91 & 0.82 & 0.88 & -0.97 \\
\midrule
\multicolumn{6}{l}{\textbf{PeopleRecognition}} \\
ResNet101 & 0.62 & 1.35 & 6.88 & 1.33 & -0.94 \\
ResNet50 & 0.68 & 1.28 & 10.07 & 1.26 & \textbf{-0.95} \\
ViT-B/16 & 0.66 & 1.22 & 2.71 & 1.23 & -0.92 \\
ViT-B/32 & \textbf{0.70} & 1.11 & 5.14 & 1.08 & -0.93 \\
ViT-L/14 (QGelu) & 0.60 & 1.20 & 0.64 & 1.09 & -0.80 \\
ViT-L/14 (baseline) & \textbf{0.70} & 1.00 & 1.00 & 1.00 & -0.90 \\
ViT-L/14-336 & 0.68 & \textbf{0.97} & \textbf{0.51} & \textbf{0.96} & -0.85 \\
\midrule
\multicolumn{6}{l}{\textbf{SatelliteImaging}} \\
ResNet101 & 0.75 & 1.49 & 2.14 & 1.40 & \textbf{-0.99} \\
ResNet50 & 0.70 & 1.65 & 1.77 & 1.50 & \textbf{-0.99} \\
ViT-B/16 & 0.80 & 1.23 & 1.31 & 1.18 & \textbf{-0.99} \\
ViT-B/32 & 0.75 & 1.40 & 1.39 & 1.27 & -0.98 \\
ViT-L/14 (QGelu) & 0.85 & \textbf{0.84} & 1.45 & \textbf{0.81} & \textbf{-0.99} \\
ViT-L/14 (baseline) & 0.84 & 1.00 & \textbf{1.00} & 1.00 & \textbf{-0.99} \\
ViT-L/14-336 & \textbf{0.87} & 0.92 & 1.37 & 0.92 & \textbf{-0.99} \\
\bottomrule
\end{tabular}
\end{table}

\section{Corruption Method Visualizations and Descriptions}
\label{appendix:corruption_visuals}

This appendix contains visual examples and the short descriptions of the predefined corruption methods used in the framework. 
The same brief descriptions are used in prompt construction (see Section~\ref{app:prompts}) and can also be found in Tables~\ref{tab:basic_corruption_methods} and~\ref{tab:advanced_corruption_methods} below.
To improve readability, the methods have been divided into two categories: basic and advanced/synthetic corruptions.

\begin{table}
    \centering
    \caption{Basic Corruption Methods and Descriptions}
    \label{tab:basic_corruption_methods}
    {\footnotesize
    \begin{tabular}{p{\linewidth}}
        \textbf{Gaussian Blur:} Smoothens the image by blurring it, reducing fine details or noise. \\
        \includegraphics[width=1.0\textwidth, alt={A sequence of images demonstrating the gaussian blur effect at increasing levels of intensity, making the base image progressively less sharp.}]{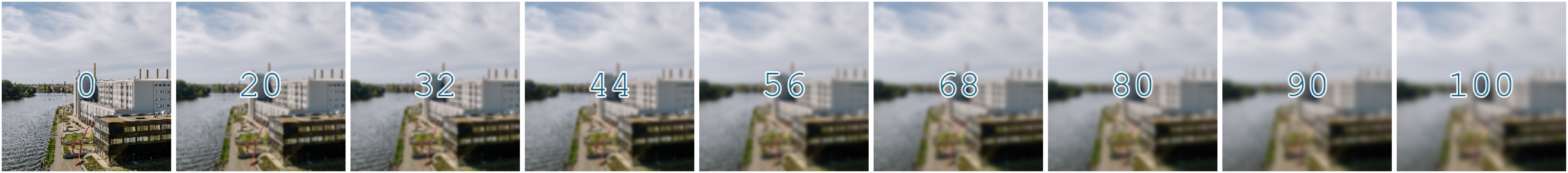} \\

        \textbf{Rotation:} Rotates the image by a specified angle, keeping its contents intact. \\
        \includegraphics[width=1.0\textwidth, alt={A sequence of images demonstrating the rotation effect, showing the base image rotated at various angles from negative to positive.}]{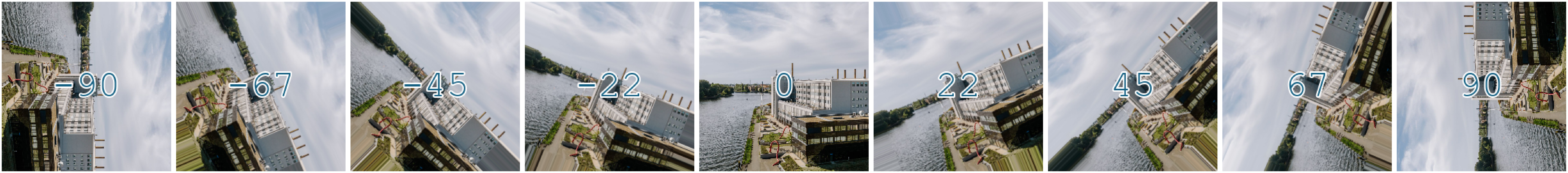} \\

        \textbf{Gaussian Noise:} Adds random, fine-grained noise following a Gaussian distribution to simulate sensor noise. \\
        \includegraphics[width=1.0\textwidth, alt={A sequence of images demonstrating gaussian noise at increasing levels, making the base image appear progressively grainier.}]{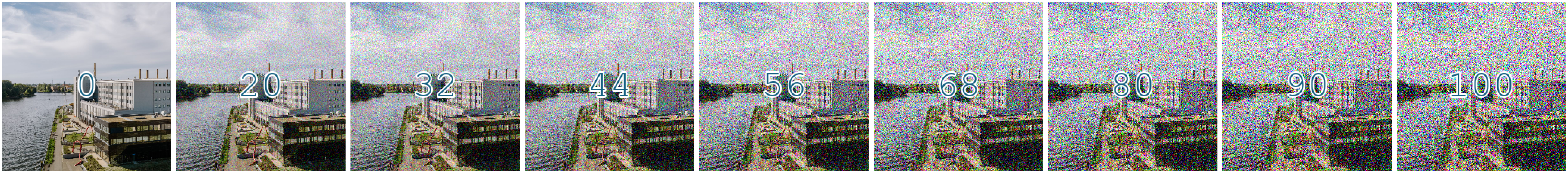} \\

        \textbf{Salt and Pepper Noise:} Adds random white and black dots to the image, mimicking noisy pixels. \\
        \includegraphics[width=1.0\textwidth, alt={A sequence of images showing salt and pepper noise applied with increasing density, covering the base image with more and more random black and white pixels.}]{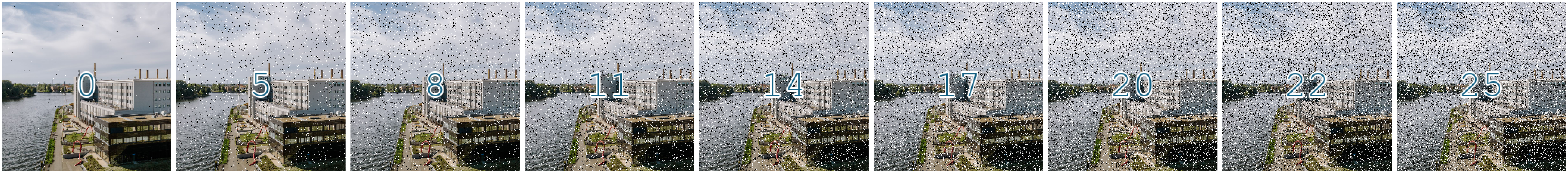} \\

        \textbf{Color Shift:} Adjusts the overall color balance of the image, shifting its tones globally. \\
        \includegraphics[width=1.0\textwidth, alt={A sequence of images demonstrating a global color shift, where the hue of the base image is progressively altered through a range of colors.}]{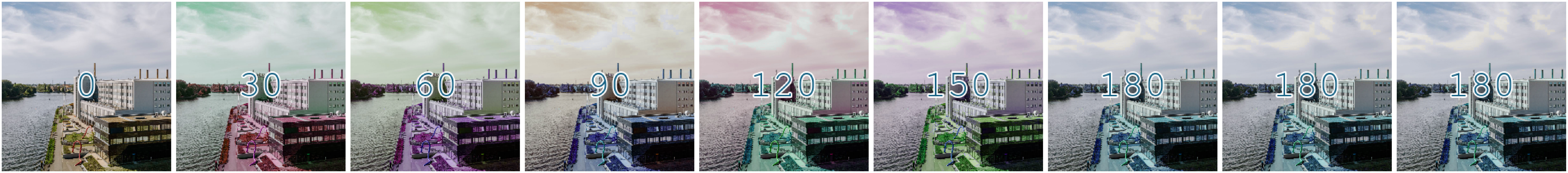} \\

        \textbf{Contrast:} Alters the difference between light and dark areas to make the image appear more or less vivid. \\
        \includegraphics[width=1.0\textwidth, alt={A sequence of images showing the effect of altering contrast, from low contrast (washed out) to high contrast (starkly defined light and dark areas).}]{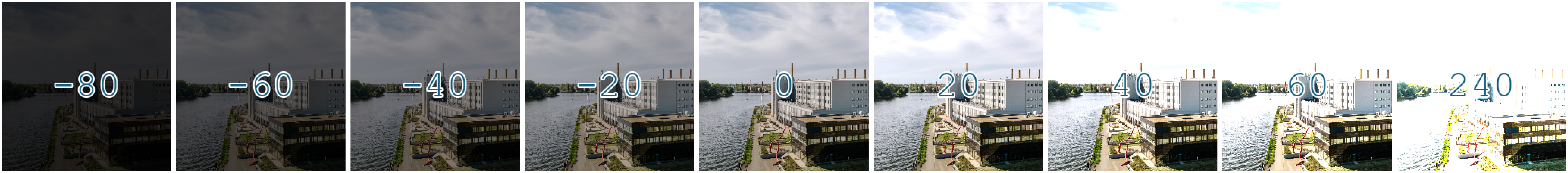} \\

        \textbf{Brightness:} Changes the overall lightness or darkness of the image. \\
        \includegraphics[width=1.0\textwidth, alt={A sequence of images demonstrating changes in brightness, with the base image progressing from very dark to very light.}]{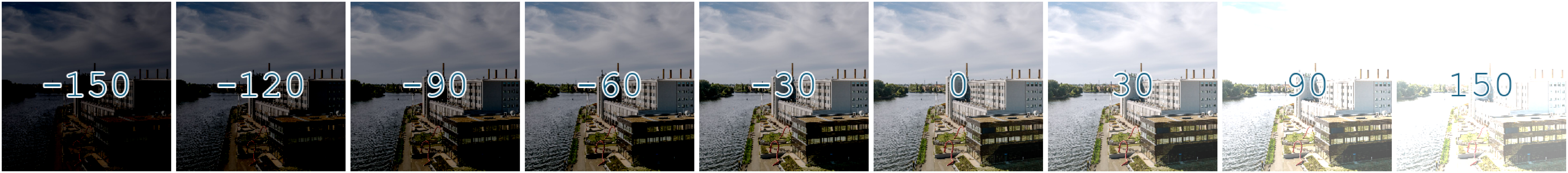} \\

        \textbf{Horizontal/Vertical Flip:} Flips the image along the horizontal/vertical axis (mirroring it). \\
        \includegraphics[width=1.0\textwidth, alt={A visual example showing the original image, followed by a horizontally flipped version, and a vertically flipped version.}]{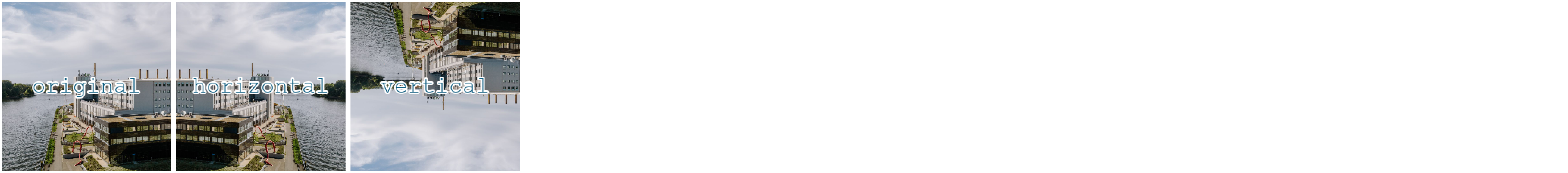}
    \end{tabular}
    }
\end{table}

\begin{table}[ht]
    \centering
    \caption{Advanced and Synthetic Corruption Methods}
    \label{tab:advanced_corruption_methods}
    {\footnotesize
    \begin{tabular}{p{\linewidth}}
        \textbf{Rain:} Adds synthetic raindrop effects or streaks to mimic rainy conditions. \\
        \includegraphics[width=1.0\textwidth, alt={A sequence of images demonstrating a synthetic rain effect. The base image is progressively covered with more and denser rain streaks.}]{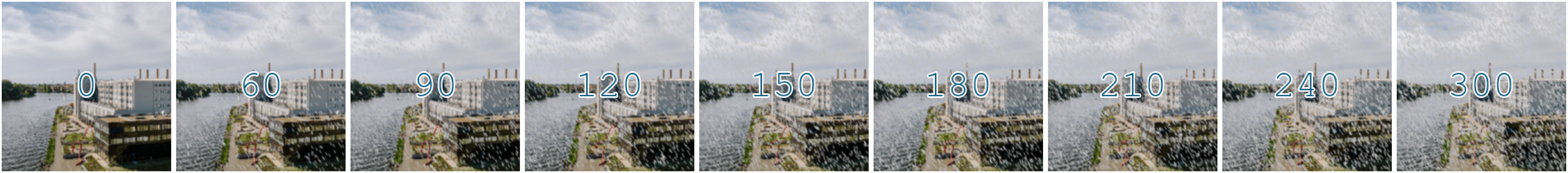} \\

        \textbf{Shadow:} Adds synthetic shadows to an image to simulate lighting conditions. \\
        \includegraphics[width=1.0\textwidth, alt={A sequence of images showing a synthetic shadow effect. A shadow is cast over the base image, growing in intensity or coverage across the frames.}]{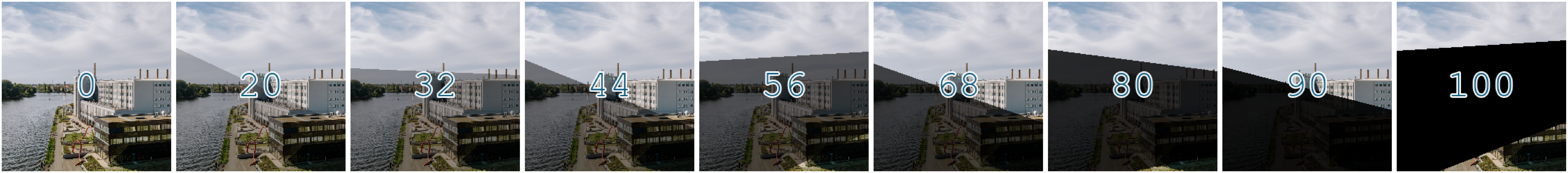} \\

        \textbf{Motion Blur:} Blurs the image to simulate movement, as if the camera or object was in motion. \\
        \includegraphics[width=1.0\textwidth, alt={A sequence of images demonstrating motion blur at increasing levels of intensity, making the base image appear as if it is moving horizontally.}]{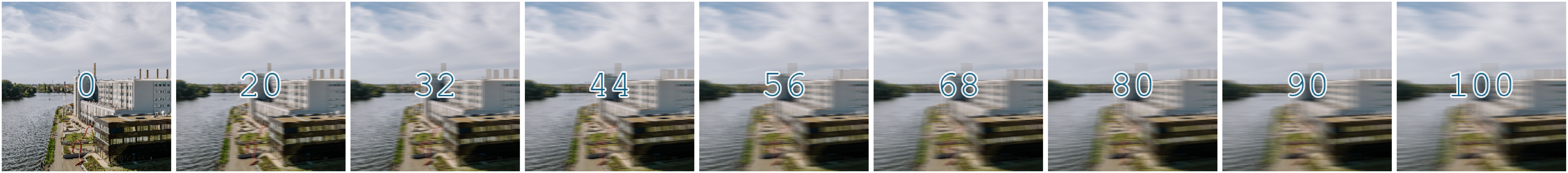} \\

        \textbf{Grid Distortion:} Distorts the image by applying a grid-like warping effect, bending specific areas. \\
        \includegraphics[width=1.0\textwidth, alt={A sequence of images showing grid distortion being applied with increasing severity, causing the base image to bulge and warp in a grid-like pattern.}]{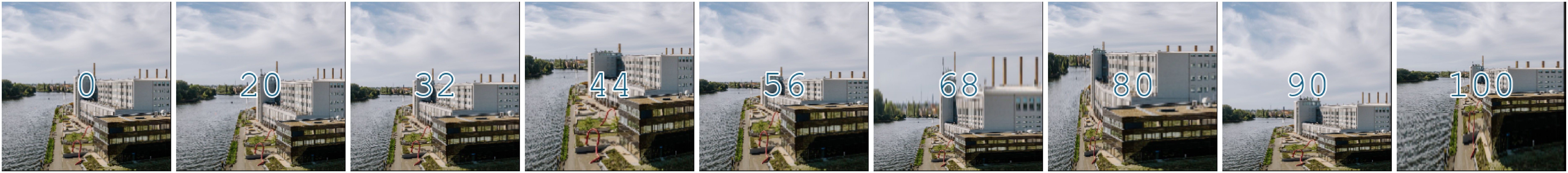} \\

        \textbf{Elastic Deformation:} Applies a rubber-sheet-like deformation to the image, bending it smoothly. \\
        \includegraphics[width=1.0\textwidth, alt={A sequence of images demonstrating elastic deformation, where the base image is smoothly stretched and distorted with increasing magnitude across the frames.}]{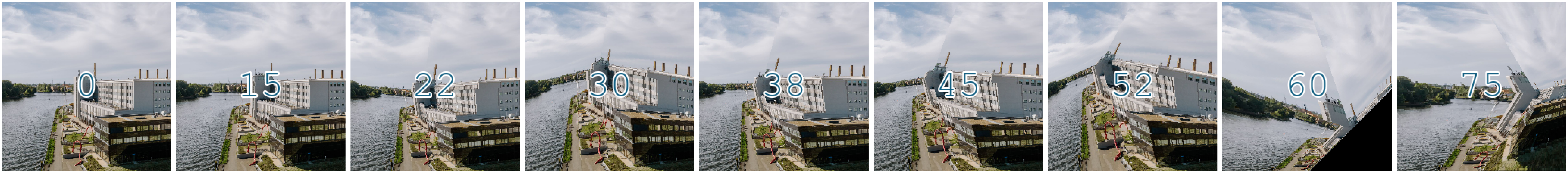} \\

        \textbf{Perspective Transform.:} Warps the image by changing its perspective, as if viewed from a different angle. \\
        \includegraphics[width=1.0\textwidth, alt={A sequence of images showing a perspective transform with increasing intensity, making the base image appear skewed and warped as if viewed from different angles.}]{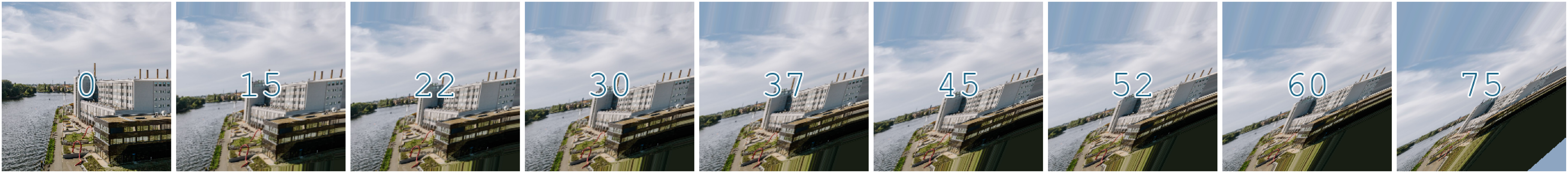} \\

        \textbf{Cloud Generator:} Overlays or generates cloud-like textures in the image, simulating an overcast sky. \\
        \includegraphics[width=1.0\textwidth, alt={A sequence of images demonstrating a synthetic cloud effect, where the sky in the base image is progressively filled with more and denser clouds.}]{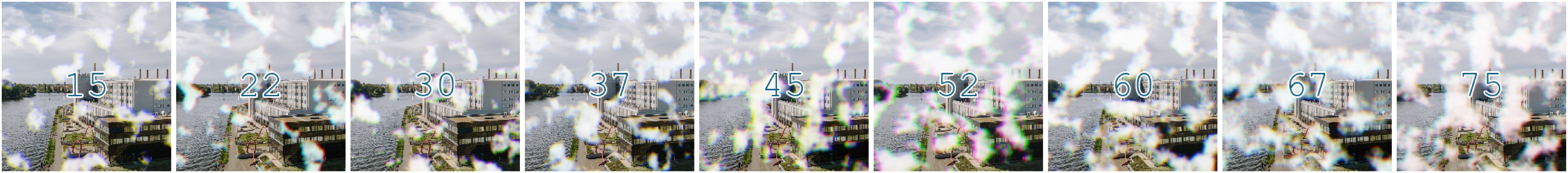}
    \end{tabular}
    }
\end{table}

\clearpage

%
\bibliographystyle{splncs04}
\bibliography{bibliography}

\begin{thebibliography}{10}
\providecommand{\url}[1]{\texttt{#1}}
\providecommand{\urlprefix}{URL }
\providecommand{\doi}[1]{https://doi.org/#1}

\bibitem{al2020dataset}
Al-Dhabyani, W., Gomaa, M., Khaled, H., Fahmy, A.: Dataset of breast ultrasound images. Data in brief  \textbf{28},  104863 (2020)

\bibitem{ali2016detection}
Ali, R.B., Ejbali, R., Zaied, M.: Detection and classification of dental caries in x-ray images using deep neural networks. In: International conference on software engineering advances (ICSEA). p.~236 (2016)

\bibitem{baby2017automatic}
Baby, M., Jereesh, A.: Automatic nerve segmentation of ultrasound images. In: 2017 International conference of electronics, communication and aerospace technology (ICECA). vol.~1, pp. 107--112. IEEE (2017)

\bibitem{bergmann2019mvtec}
Bergmann, P., Fauser, M., Sattlegger, D., Steger, C.: Mvtec ad--a comprehensive real-world dataset for unsupervised anomaly detection. In: Proceedings of the IEEE/CVF conference on computer vision and pattern recognition. pp. 9592--9600 (2019)

\bibitem{bommasani2021opportunities}
Bommasani, R., Hudson, D.A., Adeli, E., Altman, R., Arora, S., von Arx, S., Bernstein, M.S., Bohg, J., Bosselut, A., Brunskill, E., et~al.: On the opportunities and risks of foundation models. arXiv preprint arXiv:2108.07258  (2021)

\bibitem{bossard2014food}
Bossard, L., Guillaumin, M., Van~Gool, L.: Food-101--mining discriminative components with random forests. In: Computer vision--ECCV 2014: 13th European conference, zurich, Switzerland, September 6-12, 2014, proceedings, part VI 13. pp. 446--461. Springer (2014)

\bibitem{chapelle2000vicinal}
Chapelle, O., Weston, J., Bottou, L., Vapnik, V.: Vicinal risk minimization. Advances in neural information processing systems  \textbf{13} (2000)

\bibitem{cherti2023reproducible}
Cherti, M., Beaumont, R., Wightman, R., Wortsman, M., Ilharco, G., Gordon, C., Schuhmann, C., Schmidt, L., Jitsev, J.: Reproducible scaling laws for contrastive language-image learning. In: Proceedings of the IEEE/CVF Conference on Computer Vision and Pattern Recognition. pp. 2818--2829 (2023)

\bibitem{croce2020robustbench}
Croce, F., Andriushchenko, M., Sehwag, V., Debenedetti, E., Flammarion, N., Chiang, M., Mittal, P., Hein, M.: Robustbench: a standardized adversarial robustness benchmark. arXiv preprint arXiv:2010.09670  (2020)

\bibitem{cubuk2019autoaugment}
Cubuk, E.D., Zoph, B., Mane, D., Vasudevan, V., Le, Q.V.: Autoaugment: Learning augmentation strategies from data. In: Proceedings of the IEEE/CVF conference on computer vision and pattern recognition. pp. 113--123 (2019)

\bibitem{duchi2019distributionally}
Duchi, J.C., Hashimoto, T., Namkoong, H.: Distributionally robust losses against mixture covariate shifts. Under review  \textbf{2}(1) (2019)

\bibitem{fang2022data}
Fang, A., Ilharco, G., Wortsman, M., Wan, Y., Shankar, V., Dave, A., Schmidt, L.: Data determines distributional robustness in contrastive language image pre-training (clip). In: International Conference on Machine Learning. pp. 6216--6234. PMLR (2022)

\bibitem{garg2023rlsbench}
Garg, S., Erickson, N., Sharpnack, J., Smola, A., Balakrishnan, S., Lipton, Z.C.: Rlsbench: Domain adaptation under relaxed label shift. In: International Conference on Machine Learning. pp. 10879--10928. PMLR (2023)

\bibitem{geiger2012we}
Geiger, A., Lenz, P., Urtasun, R.: Are we ready for autonomous driving? the kitti vision benchmark suite. In: 2012 IEEE conference on computer vision and pattern recognition. pp. 3354--3361. IEEE (2012)

\bibitem{goodfellow2013challenges}
Goodfellow, I.J., Erhan, D., Carrier, P.L., Courville, A., Mirza, M., Hamner, B., Cukierski, W., Tang, Y., Thaler, D., Lee, D.H., et~al.: Challenges in representation learning: A report on three machine learning contests. In: Neural information processing: 20th international conference, ICONIP 2013, daegu, korea, november 3-7, 2013. Proceedings, Part III 20. pp. 117--124. Springer (2013)

\bibitem{goodfellow2015explaining}
Goodfellow, I.J., Shlens, J., Szegedy, C.: Explaining and harnessing adversarial examples. arXiv preprint arXiv:1412.6572  (2014)

\bibitem{halabi2019rsna}
Halabi, S.S., Prevedello, L.M., Kalpathy-Cramer, J., Mamonov, A.B., Bilbily, A., Cicero, M., Pan, I., Pereira, L.A., Sousa, R.T., Abdala, N., et~al.: The rsna pediatric bone age machine learning challenge. Radiology  \textbf{290}(2),  498--503 (2019)

\bibitem{han2025alignclip}
Han, Z., Luo, G., Sun, H., Li, Y., Han, B., Gong, M., Zhang, K., Liu, T.: Alignclip: navigating the misalignments for robust vision-language generalization. Machine Learning  \textbf{114}(3),  1--19 (2025)

\bibitem{hendrycks2019robustness}
Hendrycks, D., Dietterich, T.: Benchmarking neural network robustness to common corruptions and perturbations. Proceedings of the International Conference on Learning Representations  (2019)

\bibitem{jia2021scaling}
Jia, C., Yang, Y., Xia, Y., Chen, Y.T., Parekh, Z., Pham, H., Le, Q., Sung, Y.H., Li, Z., Duerig, T.: Scaling up visual and vision-language representation learning with noisy text supervision. In: International Conference on Machine Learning. pp. 4904--4916. PMLR (2021)

\bibitem{joo2023classification}
Joo, Y., Park, H.C., Lee, O.J., Yoon, C., Choi, M.H., Choi, C.: Classification of liver fibrosis from heterogeneous ultrasound image. IEEE Access  \textbf{11},  9920--9930 (2023)

\bibitem{kar2024brave}
Kar, O.F., Tonioni, A., Poklukar, P., Kulshrestha, A., Zamir, A., Tombari, F.: Brave: Broadening the visual encoding of vision-language models. In: European Conference on Computer Vision. pp. 113--132. Springer (2024)

\bibitem{koh2021wilds}
Koh, P.W., Sagawa, S., Marklund, H., Xie, S.M., Zhang, M., Balsubramani, A., Hu, W., Yasunaga, M., Phillips, R.L., Gao, I., et~al.: Wilds: A benchmark of in-the-wild distribution shifts. In: International conference on machine learning. pp. 5637--5664. PMLR (2021)

\bibitem{lee2017curated}
Lee, R.S., Gimenez, F., Hoogi, A., Miyake, K.K., Gorovoy, M., Rubin, D.L.: A curated mammography data set for use in computer-aided detection and diagnosis research. Scientific data  \textbf{4}(1), ~1--9 (2017)

\bibitem{li2022source}
Li, K., Lu, J., Zuo, H., Zhang, G.: Source-free multi-domain adaptation with generally auxiliary model training. In: 2022 International Joint Conference on Neural Networks (IJCNN). pp.~1--8. IEEE (2022)

\bibitem{lim2019fast}
Lim, S., Kim, I., Kim, T., Kim, C., Kim, S.: Fast autoaugment. Advances in neural information processing systems  \textbf{32} (2019)

\bibitem{madry2018towards}
Madry, A., Makelov, A., Schmidt, L., Tsipras, D., Vladu, A.: Towards deep learning models resistant to adversarial attacks. In: International Conference on Learning Representations (2018)

\bibitem{maharana2024enhancing}
Maharana, S.K., Zhang, B., Karlinsky, L., Feris, R., Guo, Y.: Enhancing robustness of clip to common corruptions through bimodal test-time adaptation. arXiv preprint arXiv:2412.02837  (2024)

\bibitem{openai2023gpt4}
OpenAI: Gpt-4 technical report. \url{https://openai.com/research/gpt-4} (2023), accessed: 2025-01-09

\bibitem{radford2021learning}
Radford, A., Kim, J.W., Hallacy, C., Ramesh, A., Goh, G., Agarwal, S., Sastry, G., Askell, A., Mishkin, P., Clark, J., et~al.: Learning transferable visual models from natural language supervision. In: International Conference on Machine Learning. pp. 8748--8763. PMLR (2021)

\bibitem{sarhan2024knee}
Sarhan, A.M., Gobara, M., Yasser, S., Elsayed, Z., Sherif, G., Moataz, N., Yasir, Y., Moustafa, E., Ibrahim, S., Ali, H.A.: Knee osteoporosis diagnosis based on deep learning. International Journal of Computational Intelligence Systems  \textbf{17}(1), ~241 (2024)

\bibitem{taori2020measuring}
Taori, R., Dave, A., Shankar, V., Carlini, N., Recht, B., Schmidt, L.: Measuring robustness to natural distribution shifts in image classification. Advances in Neural Information Processing Systems (NeurIPS)  \textbf{33},  18583--18599 (2020)

\bibitem{vapnik1999nature}
Vapnik, V.: The nature of statistical learning theory. Springer Science \& Business Media (1999)

\bibitem{vapnik1999overview}
Vapnik, V.N.: An overview of statistical learning theory. IEEE transactions on neural networks  \textbf{10}(5),  988--999 (1999)

\bibitem{verma2024beyond}
Verma, P., Van, M.H., Wu, X.: Beyond human vision: The role of large vision language models in microscope image analysis. arXiv preprint arXiv:2405.00876  (2024)

\bibitem{wang2024sober}
Wang, Q., Lin, Y., Chen, Y., Schmidt, L., Han, B., Zhang, T.: A sober look at the robustness of clips to spurious features. arXiv preprint arXiv:2403.11497  (2024)

\bibitem{wang2017chestx}
Wang, X., Peng, Y., Lu, L., Lu, Z., Bagheri, M., Summers, R.M.: Chestx-ray8: Hospital-scale chest x-ray database and benchmarks on weakly-supervised classification and localization of common thorax diseases. In: Proceedings of the IEEE conference on computer vision and pattern recognition. pp. 2097--2106 (2017)

\bibitem{wolf2019huggingface}
Wolf, T., Debut, L., Sanh, V., Chaumond, J., Delangue, C., Moi, A., Cistac, P., Rault, T., Louf, R., Funtowicz, M., et~al.: Huggingface's transformers: State-of-the-art natural language processing. arXiv preprint arXiv:1910.03771  (2019)

\bibitem{xia2017aid}
Xia, G.S., Hu, J., Hu, F., Shi, B., Bai, X., Zhong, Y., Zhang, L., Lu, X.: Aid: A benchmark data set for performance evaluation of aerial scene classification. IEEE Transactions on Geoscience and Remote Sensing  \textbf{55}(7),  3965--3981 (2017)

\bibitem{xu2023metaclip}
Xu, H., Xie, S., Tan, X.E., Huang, P.Y., Howes, R., Sharma, V., Li, S.W., Ghosh, G., Zettlemoyer, L., Feichtenhofer, C.: Demystifying clip data. In: 12th International Conference on Learning Representations, ICLR 2024 (2024)

\bibitem{yin2019fourier}
Yin, D., Gontijo~Lopes, R., Shlens, J., Cubuk, E.D., Gilmer, J.: A fourier perspective on model robustness in computer vision. Advances in Neural Information Processing Systems  \textbf{32} (2019)

\bibitem{zhai2023sigmoid}
Zhai, X., Mustafa, B., Kolesnikov, A., Beyer, L.: Sigmoid loss for language image pre-training. In: Proceedings of the IEEE/CVF International Conference on Computer Vision. pp. 11975--11986 (2023)

\bibitem{zheng2023preventing}
Zheng, Z., Ma, M., Wang, K., Qin, Z., Yue, X., You, Y.: Preventing zero-shot transfer degradation in continual learning of vision-language models. In: Proceedings of the IEEE/CVF International Conference on Computer Vision. pp. 19125--19136 (2023)

\end{thebibliography}

\end{document}